\documentclass[acmtog,nonacm]{acmart}

\usepackage{multirow}
\usepackage{textcomp}
\usepackage{bbold}
\usepackage{enumitem}
\usepackage[ruled]{algorithm2e}

\SetAlFnt{\small}
\SetAlCapFnt{\small}
\SetAlCapNameFnt{\small}
\SetAlCapHSkip{0pt}

\newcommand{\name}{Open-CHOIR\xspace}

\usepackage{epigraph}
\usepackage{cleveref}

\def\eg{\emph{e.g}\onedot} 
\def\ie{\emph{i.e}\onedot}

\makeatother

\makeatletter
\DeclareRobustCommand\onedot{\futurelet\@let@token\@onedot}
\def\@onedot{\ifx\@let@token.\else.\null\fi\xspace}
\makeatother

\begin{document}

\begin{teaserfigure}
    \centering
    \includegraphics[width=\linewidth]{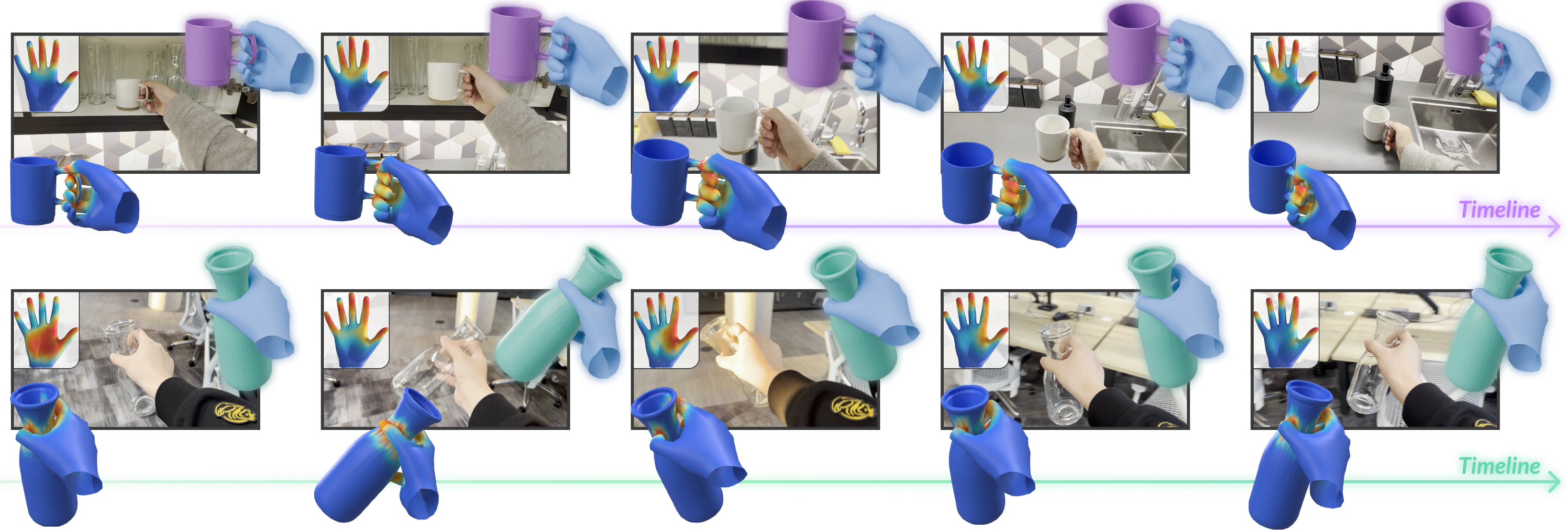}
    \caption{
        {{\color{black}From a monocular RGB video, \name reconstructs 4D hand–object interactions (HOI) (including 3D hand motion, 3D object shape, 6D pose trajectory, and contact evidence) across open-world in-the-wild scenes.
        Heatmaps reveal the HOI contact information (see supplementary for videos). }}
    }
    \label{fig:teaser}
\end{teaserfigure}

\title[Open-CHOIR: Open-World Contact-Aware 4D Hand--Object Interaction Reconstruction]{Open-CHOIR: Open-World Contact-Aware\texorpdfstring{\\}{ }4D Hand--Object Interaction Reconstruction}

\author{Hao Xu}
\affiliation{%
  \institution{The Chinese University of Hong Kong}
  \city{Hong Kong}
  \country{China}
}
\email{xuhao@cse.cuhk.edu.hk}

\author{Yilin Liu}
\affiliation{%
  \institution{University College London}
  \city{London}
  \country{United Kingdom}
}
\email{whatsevenlyl@gmail.com}

\author{Yinqiao Wang}
\affiliation{%
  \institution{The Chinese University of Hong Kong}
  \city{Hong Kong}
  \country{China}
}
\email{yqwang@cse.cuhk.edu.hk}

\author{Chi-Wing Fu}
\affiliation{%
  \institution{The Chinese University of Hong Kong}
  \city{Hong Kong}
  \country{China}
}
\email{cwfu@cse.cuhk.edu.hk}

\author{Niloy J. Mitra}
\affiliation{%
  \institution{University College London, Adobe Research}
  \city{London}
  \country{United Kingdom}
}
\email{niloym@gmail.com}

\renewcommand{\shortauthors}{Xu et al.}

\authorsaddresses{}

\begin{abstract}

We ask whether everyday open-world monocular videos can be turned into reusable 4D interaction primitives: articulated hand motion, object shape with 6D pose over time, and the when/where of contact. Such a capability would enable scalable mining of real interactions and, beyond reconstruction, support scene-aware synthesis and planning.
However, reconstructing hand-object interaction (HOI) from challenging monocular videos remains difficult: methods often assume known objects or curated scenes, and separately estimated hands and objects easily become misaligned under clutter, occlusion, and unseen object geometries.
Targeting this setting, we present {\name, an Open-world Contact-aware} HOI Reconstruction framework for a monocular camera, using contact as an explicit coupling signal between hands and objects. \name first initializes a coarse, contact-agnostic 4D HOI sequence from open-world visual priors. It then introduces a \emph{generative HOI spatial rectification} module to predict ray-depth corrections and rectify hand-object relative placement, then derive initial per-frame contact correspondences on the rectified geometry. Last, a contact-aware joint optimization with dynamically updated contact constraints enforces geometric, temporal, and contact consistency. Experiments on controlled and challenging videos show that \name improves object reconstruction, physical plausibility, and temporal consistency over state-of-the-art methods. {Code, data, and pretrained weights are available at \url{https://github.com/hxwork/CHOIR}.}
\end{abstract}


\maketitle

\section{Introduction}

\epigraph{\emph{“The world can only be grasped by action, not by contemplation. The hand is more important than the eye \dots\ The hand is the cutting edge of the mind.”}}%
{--- Jacob Bronowski, \emph{The Ascent of Man} (1973)}

Bronowski's observation frames hand-object interaction (HOI) through contact, the point at which perception becomes action. Inspired by this view, we aim to recover the underlying 4D interaction from a monocular RGB video: articulated hand motion, object shape and 6D pose over time, and when/where contact occurs. We focus on open-world manipulation videos in which the object and surrounding clutter are not known in advance, including challenging contact-bearing interaction phases with frequent occlusion and monocular depth ambiguity. If successful, such reconstruction would turn everyday videos into reusable interaction traces for mining, understanding, and scene-aware synthesis.

\begin{figure*}
    \includegraphics[width=\linewidth]{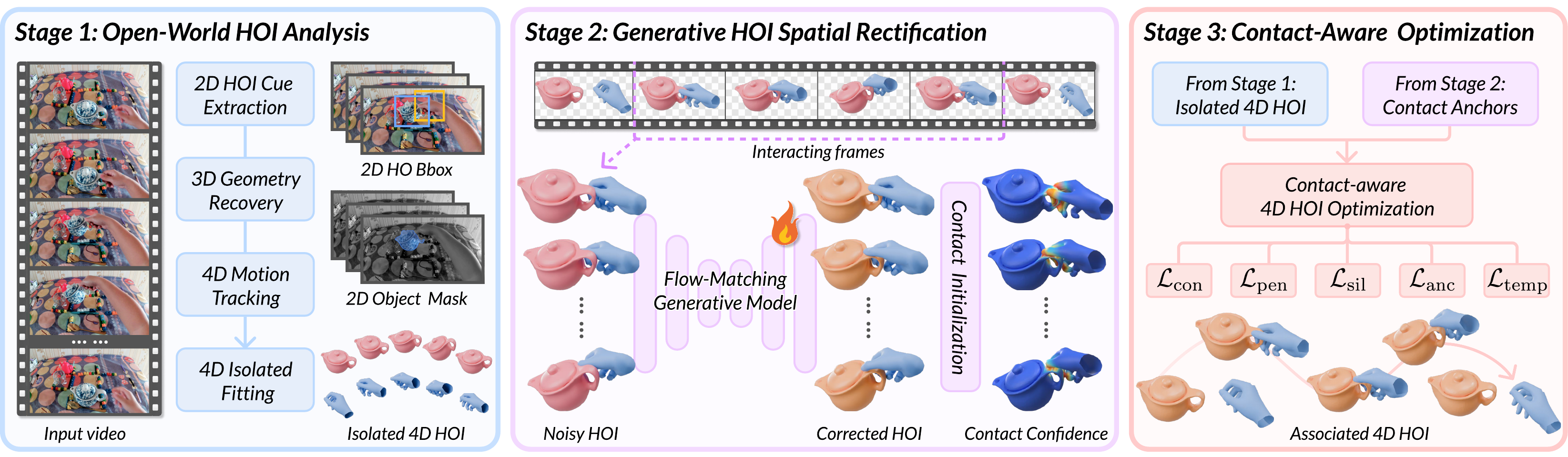}
    \vspace{-3pt}
    \caption{
    Method overview. 
    Stage~1: From a monocular video, we first obtain 2D HOI cues and initialize 3D hand/object reconstructions to form a coarse 4D HOI sequence. 
    Stage~2: {{\color{black}A flow-matching {{\color{black}generative HOI spatial rectification}} module rectifies hand-object relative placement before {{\color{black}initial}} contact correspondence construction.}} 
    Stage~3: Hierarchical contact-aware optimization refines the sequence across phases, yielding physically consistent 4D hand–object interactions.
    }
    \label{fig:pipeline}
    \vspace{-5pt}
\end{figure*}

Despite rapid progress, current HOI pipelines still struggle outside curated settings. Many methods~\cite{liu2021semi, liu2025easyhoi, wang2025magichoi} are developed and validated under restrictive assumptions, such as known object models or categories, limited clutter, mild occlusion, or carefully controlled capture. 
{{\color{black}Also, they often estimate hands and objects with weak coupling, without explicit constraints such as contact throughout reconstruction.}} 
As a result, performance degrades when the manipulated object is unseen, partially occluded, or visually ambiguous, which is precisely the regime of casual monocular videos such as Epic-Kitchens~\cite{Damen2018EPICKITCHENS} and TASTE-Rob~\cite{zhao2025taste}. 
Synthetic HOI data~\cite{hasson2019learning, wang2022dexgraspnet} and laboratory captures~\cite{GRAB2020, hampali2020honnotate, chao2021, YangCVPR2022OakInk, zhan2024oakink2} provide valuable supervision, but {{\color{black}diverse, physically grounded 4D interaction traces with object shape, object pose, hand motion, and explicit contact remain difficult to obtain at scale.}}

{{\color{black}In such cases, contact provides a sparse but strong cue for hand-object coupling, yet it becomes reliable only after the relative hand-object placement has been corrected. Thus, the difficulty is not merely to make each frame look plausible. A useful 4D HOI trace must simultaneously satisfy image evidence, metric object geometry, temporally coherent hand motion, and physically meaningful hand-object proximity. These requirements are tightly coupled: a visually plausible object mask may still imply the wrong metric scale, a smooth hand trajectory may float away from the object, and a contact cue measured before correcting relative depth can reinforce an incorrect surface correspondence.}}

{{\color{black}This motivates a staged reconstruction strategy, in which we first establish a contact-agnostic 4D sequence (Stage 1: {\em Open-World HOI Analysis\/}), then we correct the interaction-frame hand-object placement (Stage 2: {\em Generative HOI Spatial Rectification\/}) and further stabilize the coupled motion over time (Stage 3: {\em Contact-Aware Optimization\/}); see \Cref{fig:pipeline}.
Separating these roles also makes the pipeline easier to diagnose: each stage has a clear input, output, and failure mode before the final joint refinement, rather than hiding scale, contact, and temporal errors in a big optimization.}}

{{\color{black}We design \name, a contact-aware method to recover a structured 4D HOI trace from a monocular RGB video, while keeping the hand-object coupling explicit.}} {{\color{black}First, we extract 2D interaction cues and initialize a coarse, contact-agnostic 4D HOI sequence. The object is reconstructed from an anchor frame using SAM-3D-Objects~\cite{sam3dteam2025sam3d3dfyimages}, follow-tracked bidirectionally with guarded pose acceptance, and refined by isolated fitting to visually align the sequence before contact reasoning. In parallel, hand motion is initialized and temporally stabilized using recent hand reconstruction priors~\cite{pavlakos2024reconstructing, yu2025dyn}.}}

{{\color{black}Second, independently initialized hands and objects can still misalign due to wrong relative depth, causing proximity-based contact search to fail or attach the hand to an incorrect object surface. This makes contact construction a consequence of spatial correction, rather than a standalone contact-map prediction problem: before asking which surfaces touch, the hand and object must first be rectified for spatial consistency.
To do so, our generative HOI spatial rectification module uses a flow-matching prior to predict ray-depth corrections for interaction frames, refining the hand-object geometry into a contact-plausible relation, while preserving image evidence. Initial contact correspondences are then read out from the rectified geometry as hand contact vertices and barycentric object anchors. These correspondences serve as initialization evidence for optimization, rather than final reconstruction outputs.}} {{\color{black}Finally, a contact-aware joint optimization starts from this rectified initialization, periodically rebuilds a soft contact cache from the current geometry, and combines contact, penetration, silhouette, anchor, and temporal losses to refine the full 4D sequence.}}

We evaluate \name on controlled and challenging monocular videos. 
On HO3D~\cite{hampali2020honnotate}, \name improves object reconstruction and hand-object alignment over state-of-the-art baselines~\cite{fan2024hold,liu2025easyhoi, wang2025magichoi}. 
{{\color{black}For in-the-wild evaluation, we report reference-free physical and temporal metrics on a 100-video benchmark from TASTE-Rob~\cite{zhao2025taste} and self-captured videos, with supplementary multi-view video comparisons for temporal assessment. 
Results show that \name improves physical plausibility and stability while preserving image alignment (\Cref{fig:teaser}).}}

{{\color{black}In summary, our contributions are threefold. First, we introduce an open-world monocular 4D HOI reconstruction method that extracts 2D interaction cues, performs guarded object tracking, and produces a coarse contact-agnostic sequence. Second, we propose generative HOI spatial rectification to predict ray-depth corrections before constructing initial per-frame barycentric contact correspondences. Third, we develop a contact-aware joint optimization with dynamically updated contact constraints and validate it on controlled and in-the-wild videos.
}}

\section{Related Work}

\paragraph{2D Hand-Object Understanding}
Reliable 2D spatial understanding is a prerequisite for 3D/4D hand-object reconstruction.
Early works,~\eg, MediaPipe~\cite{lugaresi2019mediapipe} and 100DOH~\cite{shan2020understanding}, established valuable benchmarks for 2D hand understanding, whereas recent large-scale approaches~\cite{xu2022vitpose, cheng2023towards, potamias2025wilor} further scaled up the performance.
However, domain shifts in unconstrained videos~\cite{zhao2025taste} with cluttered backgrounds remain difficult to resolve.

\paragraph{Monocular 3D/4D Hand Reconstruction}
3D hand reconstruction has evolved from single-image mesh recovery to 4D tracking.
Early parametric methods that fit MANO~\cite{mano} to 2D observations~\cite{boukhayma20193d} have been largely superseded by transformer-based architectures. 
Notably, HaMeR~\cite{pavlakos2024reconstructing} leverages large-scale Vision Transformers for robust single-image estimation, whereas WiLoR~\cite{potamias2025wilor} refines poses for better image alignment.
Regarding videos, methods like SeqHAND~\cite{yang2020seqhand} and Deformer~\cite{fu2023deformer} enforce temporal smoothness. 
Recent approaches such as Dyn-HaMR~\cite{yu2025dyn}, HaWoR~\cite{zhang2025hawor}, and HaPTIC~\cite{ye2025predicting} advance 4D hand motion recovery by modeling camera dynamics and multi-frame priors. 
While these methods excel at isolated hand motion, they do not reconstruct manipulated object or model contact. In contrast, we tackle open-world HOI by jointly capturing both hand and object motions under explicit physical constraints.

\paragraph{Monocular 3D/4D Hand-Object Reconstruction}
Recovering HOI is challenged by unknown object geometry, severe occlusions, and coupled dynamics. 
{{\color{black}
Early learning-based methods focused on monocular hand motion capture~\cite{zhou2020monocular} or aligning pre-scanned object templates~\cite{hasson2019learning}. 
}}
To move beyond fixed templates, subsequent works adopted neural implicit representations~\cite{chen2022alignsdf, ye2022s}. 
Regarding videos, HOLD~\cite{fan2024hold} targets category-agnostic reconstruction over time, while DiffHOI~\cite{ye2023diffusion} and G-HOP~\cite{ye2024g} utilize diffusion-based refinement. 
More recently, generative priors (\eg, EasyHOI~\cite{liu2025easyhoi}, MagicHOI~\cite{wang2025magichoi}) synthesize plausible hand-object poses to handle occlusions. 
{{\color{black}
Conceptually related, BimArt~\cite{zhang2025bimart} introduces a generative prior for HOI, but focuses on the forward \textit{synthesis} of bimanual interactions.
}}

Despite these advances, recovering high-fidelity 4D HOI in an open-world setting remains non-trivial.
First, while implicit representations,~\eg, HOLD, DiffHOI, and G-HOP, offer topological flexibility, preserving geometric details from monocular inputs remains ambiguous without explicit 3D priors.
Second, generative approaches focus mainly on visual plausibility.
EasyHOI~\cite{liu2025easyhoi} excels in single-view reconstruction but does not explicitly model temporal and contact coherence required for 4D motion, while 
{{\color{black}MagicHOI~\cite{wang2025magichoi} remains sensitive to initialization errors, occlusions, and contacts in occluded regions.}}
{{\color{black}In contrast, \name targets automated 4D reconstruction with explicit contact correspondences and contact-aware joint optimization.}}
{{\color{black}
Last, while optimization-based post-processing is highly effective for enforcing validity across diverse domains~\cite{shen2025layoutrectifier, grady2021contactopt}, achieving automated, physically-grounded 4D reconstruction across an interaction lifecycle remains largely unexplored.
}}

\section{Overview}
\label{sec:overview}

\begin{figure}[t!]
    \includegraphics[width=\linewidth]{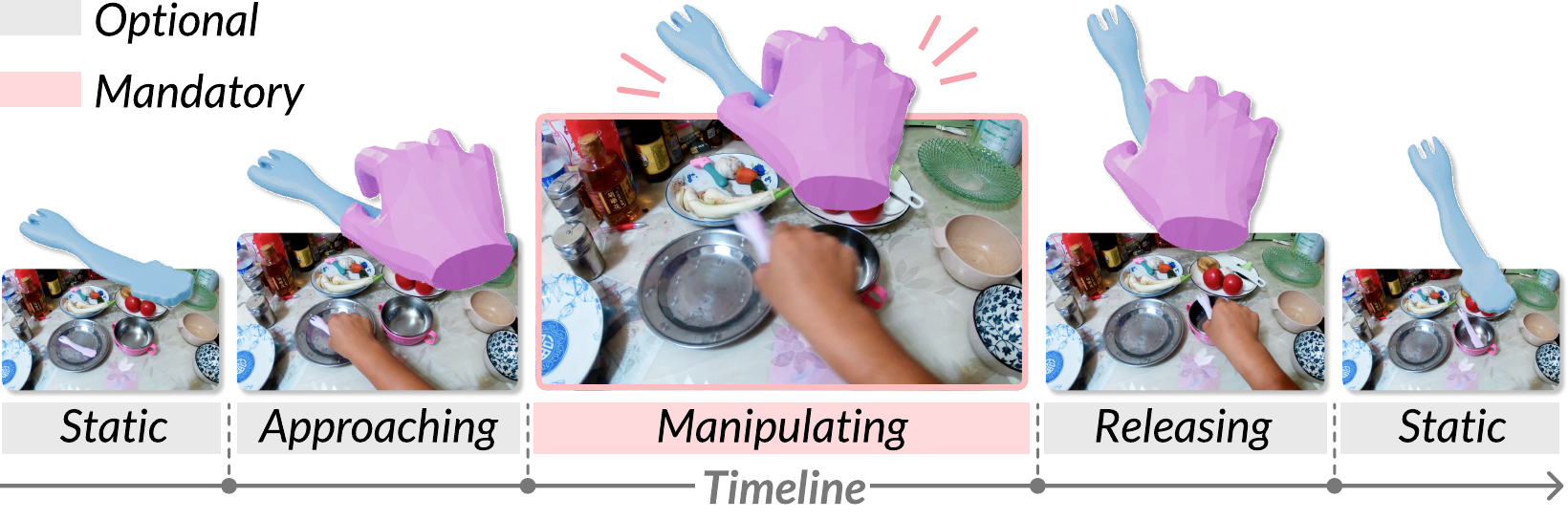}
    \vspace{-3pt}
    \caption{
    {{\color{black}Problem definition: our goal is to reconstruct 4D HOI from a monocular RGB video of a complete atomic interaction that includes (i) hand approaching a static object; (ii) grasping and manipulating the object; and (iii) putting the object back.
    Active-manipulation-only clips are also allowed.}}
    }
    \label{fig:problem_definition}
    \vspace{-5pt}
\end{figure}

\paragraph{Problem definition}
{{\color{black}
Our goal is to recover a coherent 4D representation of hand-object interactions (HOI) from a monocular RGB video captured in an open-world setting.
Specifically, the input video may capture either a complete atomic interaction, where the hand approaches a static object, grasps and manipulates it, and then returns the object to the scene (see \Cref{fig:problem_definition}), or only the active 
object manipulation phase. 
The target output includes the hand motion, object shape, object 6D pose trajectory, and contact evidence for spatially and temporally consistent 4D HOI reconstruction.
To make this goal tractable, we assume the object is rigid.
}}
%

%
%
The task presents several challenges: 

(i) \textit{2D Perception Dilemma.} \
    {{\color{black}
    Specialized interaction detectors~\cite{cheng2023towards, leonardi2024synthetic, shan2020understanding} fall short of generalizing to unseen scenarios, while general-purpose models~\cite{potamias2025wilor, karaev2025cotracker3, ravi2024sam} cannot readily comprehend HOI semantics out-of-the-box. 
    There is no off-the-shelf pipeline to robustly recognize, segment, and track the hand and specific associated object during the interaction.
    }}

(ii) \textit{3D Spatial Ambiguity.} \
    {{\color{black}
    Monocular depth and scale ambiguities, especially under hand-object occlusions, make reliable spatial alignment difficult between the hand and the object. 
    Such misalignment can make proximity-based contact search fail or attach the hand to an incorrect object surface.
    }}

(iii) \textit{4D Physical Complexity.} \
    {{\color{black}
    Extending to 4D requires maintaining physically valid contact over the frames and phases where contact is present, while also preserving image alignment in non-contact frames. 
    Since contact evidence is sparse and phase-dependent, temporal physical validity must be enforced together with penetration, silhouette, anchor, and motion priors; otherwise, the reconstruction is prone to interpenetration, floating artifacts, or temporal drift.
    }}


\paragraph{Method overview}
{{\color{black}Our key observation is that \textit{contact} offers an explicit and transferable cue for hand-object coupling during interaction.}}
{{\color{black}Though monocular perception yields noisy hand and object estimates, generative HOI spatial rectification followed by contact-aware test-time optimization refines them into a physically consistent relation.}}
\name has three stages (\Cref{fig:pipeline}):
{{\color{black}(i)~\textit{Open-world HOI analysis} extracts 2D interaction cues, recovers hand/object geometry, tracks motion, and performs coarse isolated fitting to derive a contact-agnostic 4D HOI sequence (\Cref{sec4.1});}}
{{\color{black}(ii)~\textit{Generative HOI spatial rectification} predicts ray-depth corrections with a flow-matching prior to rectify the interaction-frame hand-object placement; per-frame contact correspondences are then read out as barycentric anchors on the rectified geometry (\Cref{sec4.2}); and}}
{{\color{black}(iii)~\textit{Contact-aware joint optimization} starts from the rectified initialization and contact evidence to jointly refine hand/object trajectories with dynamically rebuilt soft contact constraints with silhouette, penetration, anchor, and temporal losses (\Cref{sec4.3}).}}

\section{Method}
\subsection{Stage 1: Open-World HOI Analysis}
\label{sec4.1}
{{\color{black}Given a monocular RGB video, Stage~1 extracts reliable image evidence and lifts it to an isolated, metrically plausible, contact-agnostic 4D HOI sequence (\Cref{fig:pipeline}). This sequence provides the coarse interaction geometry that Stage~2 rectifies before the subsequent contact-aware optimization in Stage~3. We summarize the main components below and defer details to the supplementary material.}}

\paragraph{Step 1: 2D interaction cue extraction.}
\label{sec4.1.A}
{{\color{black}For each frame, we estimate 2D interaction cues: the hand bounding box, chirality, and 2D joints; the interacting-object bounding box; and modal/amodal hand-object masks. Generic foundation models~\cite{cheng2023towards, potamias2025wilor} still suffer from chirality errors, missed detections, and weak object localization in out-of-distribution videos. We thus use a self-adaptation procedure to mine object pseudo labels around a motion-selected interaction frame, and utilize these labels to fine-tune WiLoR~\cite{potamias2025wilor} with an additional interacting-object box head, and adopt SAM~2 propagation~\cite{ravi2024sam} along with amodal video segmentation~\cite{chen2025using} to obtain temporally complete masks.}}

\paragraph{Step 2: 3D geometry recovery.}
\label{sec4.1.B}
{{\color{black}HaMeR~\cite{pavlakos2024reconstructing} initializes per-frame MANO hand pose and camera-relative translation. SAM-3D-Objects~\cite{sam3dteam2025sam3d3dfyimages} reconstructs a canonical object mesh, metric scale, and initial 6D pose from an object anchor frame. This anchor frame defaults to the first frame but can be any frame with reliable object visibility, and its reconstruction provides the fixed object geometry used for follow tracking.}}
\paragraph{Step 3: 4D motion tracking}
\label{sec4.1.C}
{{\color{black}To lift the isolated 3D estimates into a coherent reconstruction-frame sequence, VIPE~\cite{huang2025vipe} provides camera estimates and Dyn-HaMR~\cite{yu2025dyn} stabilizes the initial MANO estimates into a 4D hand trajectory. For the object, we use SAM-3D-Objects as a guarded follow tracker: starting from the anchor-frame reconstruction, we freeze the shape, scale, and translation-scale latents and update only the rotation and translation pose latents, yielding bidirectional 6D object poses over time. Outlier poses are rejected, missing poses are interpolated, and the resulting dense object pose sequence initializes the coarse isolated fitting in Step~4. The strategies are detailed in the supplementary material.}}

\paragraph{Step 4: Coarse isolated sequence fitting.}
\label{sec4.1.D}
{{\color{black}
The tracked 4D sequence obtained above is still a coarse initialization only: object poses may drift under occlusion and hand estimates may contain 2D reprojection errors or implausible articulation. We thus run an isolated fitting step before HOI spatial rectification in Stage~1, since this step still relies on image evidence and generic priors, and its goal is to produce a temporally usable coarse 4D HOI sequence for Stage~2.

{When a visible static segment is available, we refine the object pose shared within that segment and its global scale by jointly matching the tracked object silhouettes and the metric depth estimated by MoGe-2~\cite{wang2025moge2}.}

Regarding the hand, we optimize the MANO parameters with terms for preserving image alignment, monocular-depth consistency, plausible articulation, and temporal stability:
\begin{equation}
\mathcal{L}_{h}^{\mathrm{iso}}
=
\mathcal{L}_{2D}^{h}
+
\lambda_{\mathrm{depth}}\mathcal{L}_{\mathrm{depth}}^{h}
+
\lambda_{\mathrm{anat}}\mathcal{L}_{\mathrm{anat}}^{h}
+
\lambda_{\mathrm{prior}}\mathcal{L}_{\mathrm{prior}}^{h}
+
\lambda_{\mathrm{temp}}^{h}\mathcal{L}_{\mathrm{temp}}^{h},
\end{equation}
where $\mathcal{L}_{2D}^{h}$ keeps the projected joints aligned with 2D evidence; 
$\mathcal{L}_{\mathrm{depth}}^{h}$ stabilizes wrist depth using the median metric depth inside the hand mask; $\mathcal{L}_{\mathrm{anat}}^{h}$ suppresses anatomically invalid finger poses; 
$\mathcal{L}_{\mathrm{prior}}^{h}$ softly anchors the solution to the HaMeR initialization; and $\mathcal{L}_{\mathrm{temp}}^{h}$ reduces frame-to-frame jitter.
Regarding the object, direct mask-IoU optimization gives weak or unstable gradients when the rendered and target masks do not overlap. We thus use complementary silhouette terms with temporal and static consistency:
\begin{equation}
\mathcal{L}_{o}^{\mathrm{iso}}
=
\lambda_{\mathrm{rep}}\mathcal{L}_{\mathrm{rep}}
+
\lambda_{\mathrm{attr}}\mathcal{L}_{\mathrm{attr}}
+
\lambda_{\mathrm{temp}}^{o}\mathcal{L}_{\mathrm{temp}}^{o}
+
\lambda_{\mathrm{stat}}\mathcal{L}_{\mathrm{stat}}^{o},
\end{equation}
where $\mathcal{L}_{\mathrm{rep}}$ penalizes rendered pixels that spill outside the amodal mask, preventing the rendered object from leaking into background regions; 
$\mathcal{L}_{\mathrm{attr}}$ pulls projected vertices toward uncovered target-mask regions, providing useful gradients even before the masks overlap; 
$\mathcal{L}_{\mathrm{temp}}^{o}$ smooths pose changes under occlusion; and 
$\mathcal{L}_{\mathrm{stat}}^{o}$ ties static segments to a consistent object state. 
Compactly, with the rendered silhouette $\mathbf{M}$, amodal mask $\hat{\mathbf{M}}$, object vertices $\mathcal{V}$, camera projection $\Pi(\cdot)$, image  domain $\Omega$, sampled uncovered target-mask pixels $\mathcal{S}$,
\begin{equation}
\begin{aligned}
\mathcal{L}_{\mathrm{rep}}
&=
\frac{1}{|\Omega|}
\sum_{\mathbf{p}\in\Omega}
\left[\max\left(\mathbf{M}_{\mathbf{p}}-\hat{\mathbf{M}}_{\mathbf{p}},0\right)\right]^2,
\\
\mathrm{and} \
\mathcal{L}_{\mathrm{attr}}
&=
\frac{1}{|\mathcal{S}|}
\sum_{\mathbf{p}\in\mathcal{S}}
\min_{\mathbf{v}\in\mathcal{V}}
\|\mathbf{p}-\Pi(\mathbf{v})\|_2^2 .
\end{aligned}
\end{equation}
After this step, we apply a ray-scale alignment that slides the object sequence along the camera ray, so that its interaction-depth statistics match the hand, preserving the image-space silhouette fit while improving the depth initialization for {{\color{black}the Stage~2 rectification}}.
}}

\begin{figure}
    \includegraphics[width=\linewidth]{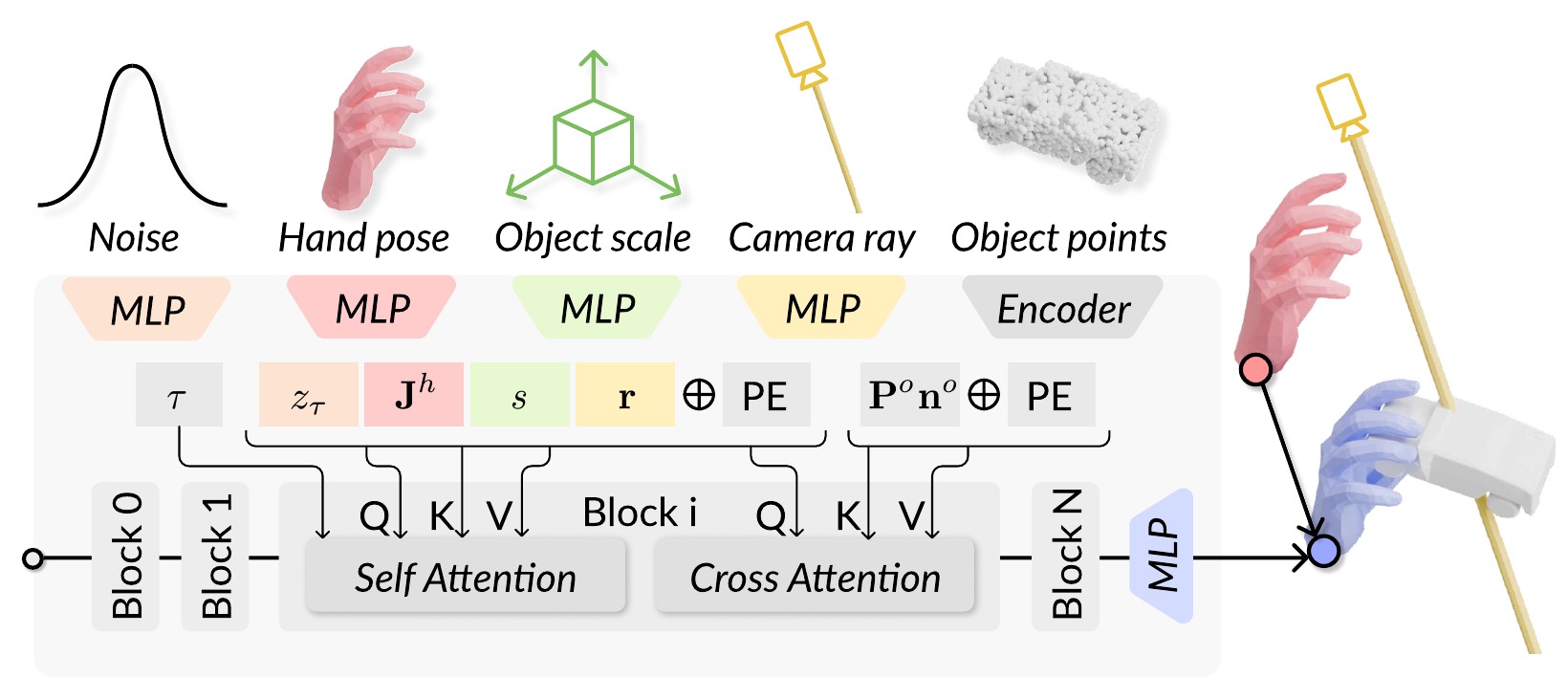}
    \vspace{-3pt}
    \caption{
    Our flow-matching-based {{\color{black}generative HOI spatial rectification}}. 
    }
    \label{fig:generative_network}
    \vspace{-5pt}
\end{figure}

\begin{figure*}[t]
    \centering
    \includegraphics[width=\linewidth]{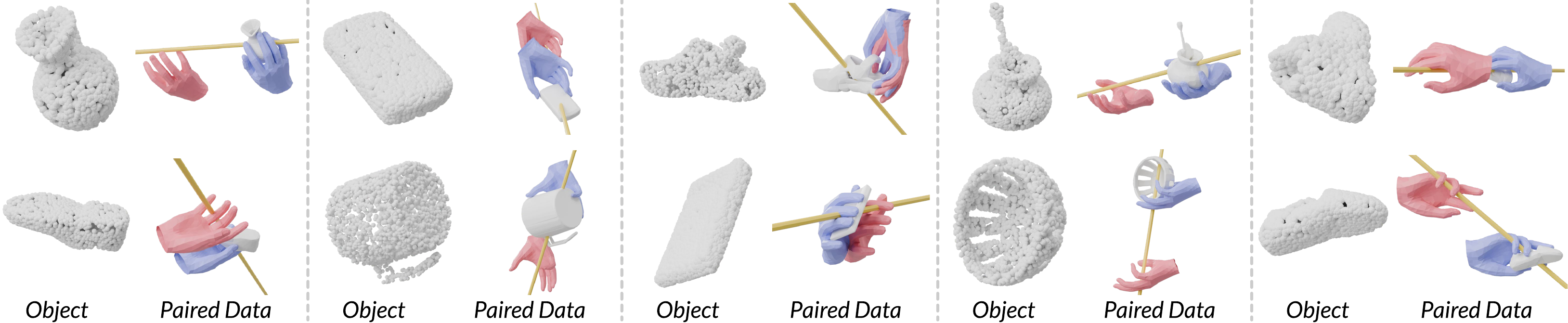}
    \vspace{-3pt}
    \caption{Examples from our GraspPair dataset. Each case comprises a posed object point cloud with 4,096 points, a source hand pose (red), a target hand pose (blue), and the camera ray direction (yellow).}
    \label{fig:grasp_pair}
    \vspace{-5pt}
\end{figure*}

\subsection{Stage 2: Generative HOI Spatial Rectification}
\label{sec4.2}
{{\color{black}The Stage~1 sequence is visually aligned but not necessarily physically valid: the hand can hover, penetrate the object, or drift under occlusion. Stage~2 first rectifies the relative hand-object placement over the interaction range, moving the noisy hand estimate into a contact-plausible alignment before any contact anchors are constructed. It then reads out {{\color{black}initial}} per-frame contact correspondences from the rectified geometry, represented by hand vertices and barycentric object anchors, rather than outputting a final optimized pose.}}

\paragraph{Modeling interaction inconsistency}
We learn a conditional model of plausible grasps using large-scale synthetic grasping data with physical filtering. 
Starting from DexGraspNet~\cite{wang2022dexgraspnet}, we simulate grasps over diverse object shapes of randomized scales and initial placements (table-top and mid-air). 
We then prune unstable grasps using a physics engine (namely, PyBullet~\cite{coumans2016pybullet}), retaining only the configurations that remain stable under gravity and perturbations,~\eg, shaking or perturbation forces. 
This yields a set of \emph{contact-valid} target grasps.

To mimic monocular ambiguity, we generate a corresponding \emph{noisy} source grasp for each simulated hand grasping pose by injecting {{\color{black}\emph{anisotropic ray-aligned}}} perturbations that emphasize depth uncertainty: large variance along the camera ray, mild in-plane noise, and anatomy-preserving perturbations in the MANO parameters. 
Each training pair thus consists of 
(i) a noisy hand estimate and viewing direction, 
(ii) the object geometry and pose context, and 
(iii) a physically valid target grasp with contact supervision.
Examples of our paired data can be found in~\Cref{fig:grasp_pair}.
\paragraph{{{\color{black}Generative HOI spatial rectification}}}
Given an object geometry and an initial (possibly misaligned) hand estimate, our goal is not to regress a single deterministic pose but to model the \emph{distribution} of physically plausible corrections with contact consistency. 
We thus formulate rectification as a conditional generative process to predict a contact-consistent hand--object alignment under occlusion.

Concretely, we represent the object by surface points (with normals) and condition the model on the current 3D hand joints, viewing direction, and object scale. We parameterize the dominant monocular failure mode (\ie, \emph{depth ambiguity}) as a 1D offset along the camera ray, and learn a flow-matching-based model~\cite{lipman2022flow} that maps a noisy depth offset to a rectified offset consistent with the object geometry, as~\Cref{fig:generative_network} shows. 
{{\color{black}Here and below, superscripts $h$ and $o$ denote hand and object quantities, respectively. Let $\mathbf{P}^{o}$ denote sampled object surface points,
$\mathbf{n}^{o}$ the associated normals, $\mathbf{J}^{h}$ the noisy hand joints, $\mathbf{r}$ the viewing ray, and $s$ the object scale. The model predicts a scalar correction along viewing ray $\mathbf{r}$, because this depth direction captures the dominant monocular ambiguity while preserving the image-space hand configuration. With flow-matching model $v_\theta$, time $\tau\in[0,1]$, and state $z_\tau\in\mathbb{R}^{1}$, {{\color{black}where $z_\tau$ is the noisy intermediate ray-offset state between the source offset and the rectified target offset,}} the rectification velocity is}}
\begin{equation}
\frac{\mathrm{d}z_\tau}{\mathrm{d}\tau}
=
v_\theta(z_\tau,\tau \mid \mathbf{J}^h,\mathbf{r},s,\mathbf{P}^{o},\mathbf{n}^{o}).
\end{equation}
{At frame $t$, we use the ray from the camera center $\mathbf{c}_t$ to the current MANO root (global hand translation) $\mathbf{p}^{h}_{t,\mathrm{root}}$: $\mathbf{r}_t=(\mathbf{p}^{h}_{t,\mathrm{root}}-\mathbf{c}_t)/\|\mathbf{p}^{h}_{t,\mathrm{root}}-\mathbf{c}_t\|_2$. The sampled correction translates the entire hand as $\mathbf{p}^{h\prime}_{t,\mathrm{root}}=\mathbf{p}^{h}_{t,\mathrm{root}}+\Delta z_t\mathbf{r}_t$, preserving its articulation and image-space configuration.}
{{\color{black}Sampling the conditional flow places the noisy hand near a contact-plausible object surface, from which we compute {{\color{black}initial contact anchors}}. {We implement $v_\theta$ by encoding the normalized object point cloud into coordinate-sensitive geometric features and fusing them with tokenized representations of the flow state, hand joints, object scale, and viewing ray to predict the rectification velocity; details are provided in the supplementary material.}}}
\begin{figure*}[t]
    \centering
    \includegraphics[width=\linewidth]{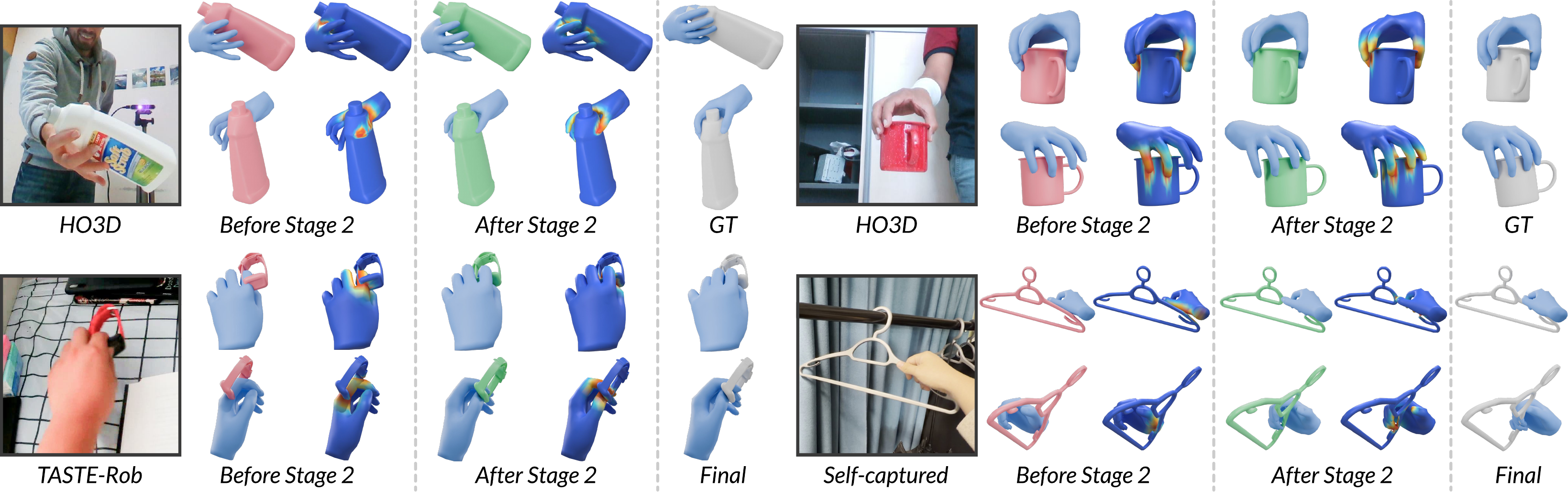}
    \vspace{-3pt}
    \caption{{{\color{black}Visualization of Stage~2 rectification and contact correspondences on HO3D, TASTE-Rob, and self-captured videos. We compare the reconstruction before and after Stage~2, together with ground truths when available or the final optimized result otherwise.}}}
    \label{fig:contact}
    \vspace{-5pt}
\end{figure*}

\paragraph{Contact correspondence establishment}
{{\color{black}
For each interaction frame, we align the object mesh to canonical space via Procrustes registration and transform the rectified hand accordingly. For candidate MANO contact vertices on the fingertips, finger pads, and palm, we query densely sampled object-surface points and retain neighbors satisfying proximity and normal-compatibility criteria.
{In practice, we use a fixed seed to sample $\mathcal{P}^{o}=\{\mathbf{p}^{o}_{j}\}_{j=1}^{10{,}000}$ and query $50$ neighbors for each transformed contact vertex $\widetilde{\mathbf{v}}^{h}_{t,i}$. Here, $j$ indexes these surface samples, and $\mathcal{N}_{t,i}$ contains the $50$ nearest such indices. Let $\rho_{t,i,j}=\|\mathbf{p}^{o}_{j}-\widetilde{\mathbf{v}}^{h}_{t,i}\|_2$ and $\alpha_{t,i,j}=\arccos\!\bigl((\widetilde{\mathbf{n}}^{h}_{t,i})^{\!\top}(\mathbf{p}^{o}_{j}-\widetilde{\mathbf{v}}^{h}_{t,i})/\rho_{t,i,j}\bigr)$, where $\widetilde{\mathbf{n}}^{h}_{t,i}$ is the contact-facing hand normal. The accepted neighbors are}
\begin{equation}
{
\begin{aligned}
\mathcal{N}_{t,i}&=\operatorname{KNN}_{50}(\widetilde{\mathbf{v}}^{h}_{t,i},\mathcal{P}^{o})\quad\mathrm{and}\\
\mathcal{A}_{t,i}&=\{j\in\mathcal{N}_{t,i}\mid
\rho_{t,i,j}<2\,\mathrm{cm},\ \alpha_{t,i,j}<60^\circ\}.
\end{aligned}
}
\end{equation}

{Each accepted point $\mathbf{p}^{o}_{j}$ is stored as an initial per-frame correspondence $(i,f_{t,i,j},\mathbf{b}_{t,i,j})$, where $f_{t,i,j}$ is its mesh face and $\mathbf{b}_{t,i,j}$ its barycentric coordinates. With face vertices $\mathbf{q}^{o}_{\ell}$ at frame $t$, its anchor is $\mathbf{a}^{o}_{t,i,j}=\sum_{\ell=1}^{3}b_{\ell}\mathbf{q}^{o}_{\ell}$.} These {{\color{black}initial}} anchors provide contact evidence for Stage~3, where the soft contact cache is rebuilt and stabilized during optimization.
}}

\subsection{Stage 3: Contact-aware HOI Optimization}
\label{sec4.3}
{{\color{black}
Stage~1 provides a visually-aligned but not necessarily physically valid output, whereas Stage~2 provides a rectified interaction sequence with contact correspondences but not final poses. 
Stage~3 then jointly optimizes the hand and object trajectories from the rectified initialization, using the Stage~2 correspondences as contact evidence and image-alignment and temporal priors from Stage~1.

For each clip, we consider up to five phases--pre-static, approach, 
manipulation, release, and post-static--although many clips contain only a subset of them (\Cref{fig:problem_definition}).
Contact losses are active on frames where contact correspondences are available; silhouette, penetration, anchor, and temporal terms provide context across the whole sequence. 
{We identify these phases from temporal changes in tracked object motion and amodal-mask overlap, and apply phase-specific objectives only when the corresponding phase is present.}
The contact-aware optimization minimizes five weighted objectives:
\begin{equation}
\begin{aligned}
\mathcal{L}_{\mathrm{HOI}}
&=
\lambda_{\mathrm{con}}\mathcal{L}_{\mathrm{con}}
+
\lambda_{\mathrm{pen}}\mathcal{L}_{\mathrm{pen}}
+
\lambda_{\mathrm{sil}}\mathcal{L}_{\mathrm{sil}}
+
\lambda_{\mathrm{anc}}\mathcal{L}_{\mathrm{anc}}
+
\lambda_{\mathrm{temp}}\mathcal{L}_{\mathrm{temp}}.
\end{aligned}
\end{equation}
$\mathcal{L}_{\mathrm{con}}$ pulls active contact vertices toward object-surface anchors; 
$\mathcal{L}_{\mathrm{pen}}$ penalizes interpenetration; 
$\mathcal{L}_{\mathrm{sil}}$ aligns rendered silhouettes to amodal masks; 
$\mathcal{L}_{\mathrm{anc}}$ collects object/hand anchors and reliable 2D/anatomical evidence; and 
$\mathcal{L}_{\mathrm{temp}}$ enforces smooth motion. 

\paragraph{Contact and penetration constraints.}
Starting from the Stage~2 rectified interaction frames, the optimizer periodically rebuilds a soft top-$K$ correspondence cache from the current hand-object geometry. Each active MANO contact vertex is associated with object-surface anchors represented by face ids and barycentric coordinates. A representative term in $\mathcal{L}_{\mathrm{con}}$ is the soft correspondence loss
\begin{equation}
\mathcal{L}_{\mathrm{contact}}
=
\frac{
\sum_{t,i}
c_{t,i}
\sum_{k=1}^{K}
\omega_{t,i,k}
\|\mathbf{v}^{h}_{t,i}-\mathbf{a}^{o}_{t,i,k}\|_2^2
}{
\sum_{t,i}c_{t,i}
},
\end{equation}
where 
$\mathbf{v}^{h}_{t,i}$ is hand contact vertex $i$ at frame $t$; 
$c_{t,i}$ is its contact confidence/gate, computed from the rebuilt cache using proximity, normal compatibility, and temporal memory; 
$\omega_{t,i,k}$ is a normalized softmax weight over the top-$K$ candidate object anchors, computed from anchor distances; and 
$\mathbf{a}^{o}_{t,i,k}$ is the dynamically rebuilt barycentric object anchor.
{At each cache rebuild, we retain the $K=8$ closest anchors within $5\,\mathrm{cm}$ and the same $60^\circ$ normal cone, using distance-based softmax weights with temperature $\sigma=0.01$. The inner sum aggregates anchors per contact vertex, and the confidence-normalized outer sum aggregates active vertices and frames; temporal memory in $c_{t,i}$ downweights unstable correspondences between rebuilds.}
If no active contacts are available in a clip, we skip $\mathcal{L}_{\mathrm{contact}}$.
In parallel, a one-sided penetration term, evaluated over all optimized frames, pushes hand vertices that lie inside the object back toward the object surface. 
Let {{\color{black}$\mathcal{V}^{h}$ $= \{\mathbf{v}^{h}\}$ be the set of hand mesh vertices,}} $d_o(\mathbf{v}^{h})$ be the
signed penetration depth that is positive inside the object and non-positive outside, and 
$\epsilon_{\mathrm{pen}}$ be a small tolerance. We use
\begin{equation}
\mathcal{L}_{\mathrm{pen}}
=
\frac{1}{|\mathcal{V}^{h}|}
\sum_{\mathbf{v}^{h}}
\max\left(d_o(\mathbf{v}^{h})-\epsilon_{\mathrm{pen}},\;0\right).
\end{equation}
This stage keeps the object pose trainable, where the silhouette and anchor groups prevent contact constraints from moving the object away from the image evidence.

\paragraph{Silhouette, anchor, and temporal stabilization.}
The remaining loss families stabilize the optimization when contact evidence is sparse or phase-dependent. We use the following compact definitions:
\begin{equation}
\begin{aligned}
\mathcal{L}_{\mathrm{sil}}
&=
\frac{1}{T}\sum_t
\ell_{\mathrm{mask}}\!\left(
\mathcal{R}(\mathbf{O}_t),
\hat{\mathbf{M}}^o_t
\right),\\
\mathcal{L}_{\mathrm{anc}}
&=
\mathcal{L}_{2D}^{h}
+
\mathcal{L}_{\mathrm{anat}}^{h}
+
\mathcal{L}_{\mathrm{pose}}^{h}
+
\mathcal{L}_{\mathrm{pose}}^{o},
\\
\mathrm{and} \
\mathcal{L}_{\mathrm{temp}}
&=
\frac{1}{T}\sum_t
\left(
\|\Delta \mathbf{x}_t\|_2^2
+
\|\Delta^2 \mathbf{x}_t\|_2^2
\right),
\end{aligned}
\end{equation}
where $T$ is the number of optimized frames; $\mathcal{R}(\mathbf{O}_t)$ is the rendered object silhouette from object state $\mathbf{O}_t$; $\hat{\mathbf{M}}^o_t$ is the amodal object mask; $\ell_{\mathrm{mask}}$ is the silhouette discrepancy implemented as the MSE between the rendered and target amodal masks; 
$\mathcal{L}_{2D}^{h}$, $\mathcal{L}_{\mathrm{anat}}^{h}$, $\mathcal{L}_{\mathrm{pose}}^{h}$, $\mathcal{L}_{\mathrm{pose}}^{o}$ represent anchor terms on 2D hand joint, 3D hand anatomy, hand pose, and object pose, respectively;
and $\mathbf{x}_t$ collects the optimized hand pose, wrist motion, and object pose, with $\Delta$ and $\Delta^2$ denoting the first- and second-order temporal finite differences.
}}

\section{Experiments}

\paragraph{Baselines}
We compare \name against five representative state-of-the-art template-free HOI reconstruction baselines: iHOI~\cite{ye2022s}, DiffHOI~\cite{ye2023diffusion}, HOLD~\cite{fan2024hold}, EasyHOI~\cite{liu2025easyhoi}, and MagicHOI~\cite{wang2025magichoi}.

\paragraph{Training dataset}
To train our {{\color{black}generative HOI spatial rectification}} module (\Cref{sec4.2}), we build \textit{GraspPair} from DexGraspNet~\cite{wang2022dexgraspnet}, a large-scale synthetic dataset containing $\sim$500k paired hand-object grasps over 5,355 object instances from 133 categories.
Each object is paired with 200 physically valid grasp configurations on average, providing diverse compatible 3D hand poses.
These pairs supervise the flow-matching model to learn plausible hand-object spatial relationships that generalize to unseen objects.
\paragraph{Evaluation benchmarks}
{{\color{black}
We evaluate metric accuracy and open-world robustness using data from three sources. 
(i) \textit{HO3D}~\cite{hampali2020honnotate} provides lab-captured videos with ground-truth 3D annotations; following standard protocols~\cite{fan2024hold, wang2025magichoi}, we use its 14-sequence subset for quantitative metric evaluation. 
(ii) \textit{TASTE-Rob}~\cite{zhao2025taste} provides diverse monocular sequences that feature rigid objects in cluttered backgrounds and long-tail scenarios, from which we select {{\color{black}70}} challenging single-hand clips. 
(iii) \textit{Self-captured videos} comprise {{\color{black}30}} casually recorded clips of diverse objects, scenes, and motions to test unconstrained generalizability. 
To assess \textit{in-the-wild} performance, where 3D ground truths are not available, we curate a comprehensive benchmark combining TASTE-Rob and self-captured videos.
For each baseline, we compute metrics only on sequences for which at least one third of frames produce a reconstruction.
As the baseline outputs contain outliers, we report mean, standard deviation, and median for the in-the-wild comparison.
Details are provided in the supplementary material.
}}

\paragraph{Metrics}
We evaluate hand pose, object shape, and hand–object alignment quality. Following~\cite{fan2024hold, ye2023diffusion, wang2025magichoi}, we report root-relative MPJPE (mm) for hand joints and Chamfer Distance (CD, cm) for object reconstruction. 
For the remaining HO3D object-centric metrics, we follow the MagicHOI evaluation protocol~\cite{wang2025magichoi} to compute F-scores at $5\,$mm (F5) and $10\,$mm (F10) for local shape detail. 
For hand-relative alignment, we translate the predicted object mesh by the estimated hand-root, then compute hand-relative Chamfer distance ($\mathrm{CD}_h$). 
Lastly, we report Relative Scale (RS), where $s_{\mathrm{ICP}}$ is the ICP scale to align the predicted mesh to ground truth. 
$\mathrm{RS}=\frac{1}{s_{\mathrm{ICP}}}-1$ if $s_{\mathrm{ICP}}<1$ and $\mathrm{RS}=1-\frac{1}{s_{\mathrm{ICP}}}$ if $s_{\mathrm{ICP}}\ge1$; $\mathrm{RS}=0$ indicates correct physical size.

{{\color{black} 
As the in-the-wild benchmark lacks 3D ground truths, we use the following reference-free metrics.
(i) \textit{Physical plausibility}: We calculate the hand--object closest distance (H-O Dist.) and penetration volume ratio (Pen. Ratio) to evaluate the amount of floating and intersection artifacts. 
(ii) \textit{Video alignment}: We compute the Mask Intersection over Union (mIoU) between the rendered 3D silhouettes and tracked 2D masks to evaluate the spatial fidelity. 
(iii) \textit{Temporal smoothness}: We report acceleration in $\text{cm/frame}^2$. $\text{Acc}_h$ is the average second-order finite difference of 3D hand joints across frames.
$\text{Acc}_o$ is computed from object centers to accommodate baselines whose reconstructed object geometry varies across frames.
}}

\begin{table}[t]
    \centering
    \caption{
        {{\color{black}Quantitative comparison on HO3D using ground-truth reconstruction metrics (top) and reference-free metrics (bottom). Best results are bold.}}
        }
    \vspace{-3pt}
    \label{tab:quantitative_comparison}
    \begingroup
        \footnotesize
        \begin{tabular*}{\linewidth}{@{\extracolsep{\fill}}rcccccc@{}}
            \toprule
            \multirow{2}{*}{Method} & CD & F5 & F10 & MPJPE & CD$_h$ & RS \\
            \cmidrule(lr){2-7}
             & [cm] $\downarrow$ & [\%] $\uparrow$ & [\%] $\uparrow$ & [mm] $\downarrow$ & [cm] $\downarrow$ & $\downarrow$ \\
            \midrule
            iHOI       & 2.37 & 35.78 & 62.11 & 27.75 & 25.45 & 0.17 \\
            DiffHOI   & 2.30 & 39.59 & 64.49 & 16.02 & 33.33 & 0.13 \\
            EasyHOI    & 1.86 & 46.10 & 70.92 & 16.69 & 19.55 & 0.28 \\
            HOLD       & 1.31 & 57.20 & 80.23 & 30.79 & 21.28 & 0.62 \\
            MagicHOI   & 0.87 & 69.72 & 92.15 & \textbf{4.62} & \textbf{2.39} & 0.11 \\
            \midrule
            \textbf{\name (ours)} & \textbf{0.77} & \textbf{72.33} & \textbf{96.03} & 5.55 & 2.41 & \textbf{0.10} \\
            \bottomrule
        \end{tabular*}
    
    \par\smallskip
        \begin{tabular*}{\linewidth}{@{\extracolsep{\fill}}rccccc@{}}
            \toprule
            \multirow{2}{*}{Method} & mIoU & Pen. Ratio & H-O Dist. & Acc$_h$ & Acc$_o$ \\
            \cmidrule(lr){2-6}
            & [\%] $\uparrow$ & [\%] $\downarrow$ & [cm] $\downarrow$ & [cm/fr$^2$] $\downarrow$ & [cm/fr$^2$] $\downarrow$ \\
            \midrule
            HOLD     & 75.01 & 17.72 & 0.10 & 2.07 & \textbf{1.03} \\
            MagicHOI & 70.57 & 14.11 & 0.36 & 9.51 & 7.51 \\
            \textbf{\name (ours)} & \textbf{85.87} & \textbf{4.58} & \textbf{0.06} & \textbf{1.75} & 1.08 \\
            \bottomrule
        \end{tabular*}
    \endgroup
    \vspace{-5pt}
\end{table}

\begin{table*}[t]
    \centering
    \caption{{Quantitative comparison on the in-the-wild benchmark. HOLD-valid and MagicHOI-valid denote subsets where the corresponding baseline can successfully reconstruct at least one third of the frames. Within each subset, the baseline and \name are evaluated on the same videos; results are comparable only within that set. The final row reports \name on all 100 videos. Std. and Med. denote standard deviation and median, respectively.}}
    \vspace{-3pt}
    \label{tab:in_the_wild_comparison}
    \begingroup
        \footnotesize
        \setlength\tabcolsep{1.5pt}
        \begin{tabular*}{\linewidth}{@{\extracolsep{\fill}}llccccccccccccccc@{}}
            \toprule
            \multirow{3}{*}{Method}
            & \multirow{3}{*}{Evaluation set}
            & \multicolumn{3}{c}{mIoU [\%] $\uparrow$}
            & \multicolumn{3}{c}{Pen. Ratio [\%] $\downarrow$}
            & \multicolumn{3}{c}{H-O [cm] $\downarrow$}
            & \multicolumn{3}{c}{Acc$_h$ [cm/fr$^2$] $\downarrow$}
            & \multicolumn{3}{c}{Acc$_o$ [cm/fr$^2$] $\downarrow$} \\
            \cmidrule(lr){3-5}
            \cmidrule(lr){6-8}
            \cmidrule(lr){9-11}
            \cmidrule(lr){12-14}
            \cmidrule(lr){15-17}
            & & Mean & Std. & Med. & Mean & Std. & Med. & Mean & Std. & Med. & Mean & Std. & Med. & Mean & Std. & Med. \\
            \midrule
            HOLD & HOLD-valid (30) & 27.19 & 29.08 & 10.18 & 1.08 & 1.81 & 0.06 & 194.71 & 528.52 & 5.62 & 36.03 & 114.39 & 4.15 & 32.61 & 73.61 & 3.11 \\
            \name (ours) & HOLD-valid (30) & 78.85 & 11.40 & 83.43 & 6.12 & 5.28 & 4.26 & 0.09 & 0.08 & 0.06 & 0.43 & 0.25 & 0.41 & 0.12 & 0.08 & 0.10 \\
            \midrule
            MagicHOI & MagicHOI-valid (22) & 23.20 & 21.88 & 16.59 & 3.95 & 6.41 & 1.07 & 6.82 & 10.53 & 2.20 & 3.71 & 5.15 & 2.02 & 2.70 & 5.10 & 1.18 \\
            \name (ours) & MagicHOI-valid (22) & 77.89 & 11.62 & 80.56 & 6.51 & 5.99 & 4.26 & 0.11 & 0.08 & 0.06 & 0.36 & 0.21 & 0.36 & 0.11 & 0.08 & 0.10 \\
            \midrule
            \name (ours) & All videos (100) & 76.65 & 11.93 & 79.13 & 5.81 & 4.92 & 4.00 & 0.10 & 0.13 & 0.06 & 0.42 & 0.26 & 0.41 & 0.12 & 0.08 & 0.10 \\
            \bottomrule
        \end{tabular*}
    \endgroup
    \vspace{-5pt}
\end{table*}

\paragraph{Implementation details}
We implement our method using PyTorch~\cite{pytorch19} and adopt the Adam optimizer~\cite{kingma2014adam} for both training the feed-forward model (Stage~2) and optimizing the hand/object poses in Stage~1 and Stage~3.
{{\color{black}Please refer to the supplementary material.}}

\paragraph{Quantitative comparison}
{{\color{black}\Cref{tab:quantitative_comparison} reports results on HO3D for both ground-truth reconstruction metrics and reference-free metrics. 
\name achieves the best object CD, F-scores, and scale error, indicating more accurate object shape and 6D pose recovery. 
MagicHOI~\cite{wang2025magichoi} obtains a lower MPJPE by freezing the HaMeR~\cite{pavlakos2024reconstructing} hand estimate, while our method uses HaMeR only as initialization and jointly optimizes hand and object alignment. 
The lower penetration ratio, smaller hand-object distance, and smoother hand motion further show the benefit of contact-aware optimization beyond object reconstruction.
%

\Cref{tab:in_the_wild_comparison} reports mean, standard deviation, and median for in-the-wild videos with no 3D ground truths. 
HOLD and MagicHOI produce usable reconstructions on 30 and 22 out of 100 videos, respectively; most other sequences fail due to COLMAP-based tracking breakdowns.
For a fair comparison, we evaluate \name on the same baseline-valid subsets (HOLD-valid or MagicHOI-valid) and on the full benchmark.
Even on their valid subsets, the baselines show low median mIoU and large hand-object distances, while \name maintains stronger alignment and smoother hand/object trajectories for both subset and full evaluations.}}

\begin{figure}[t]
    \centering
    \includegraphics[width=\linewidth]{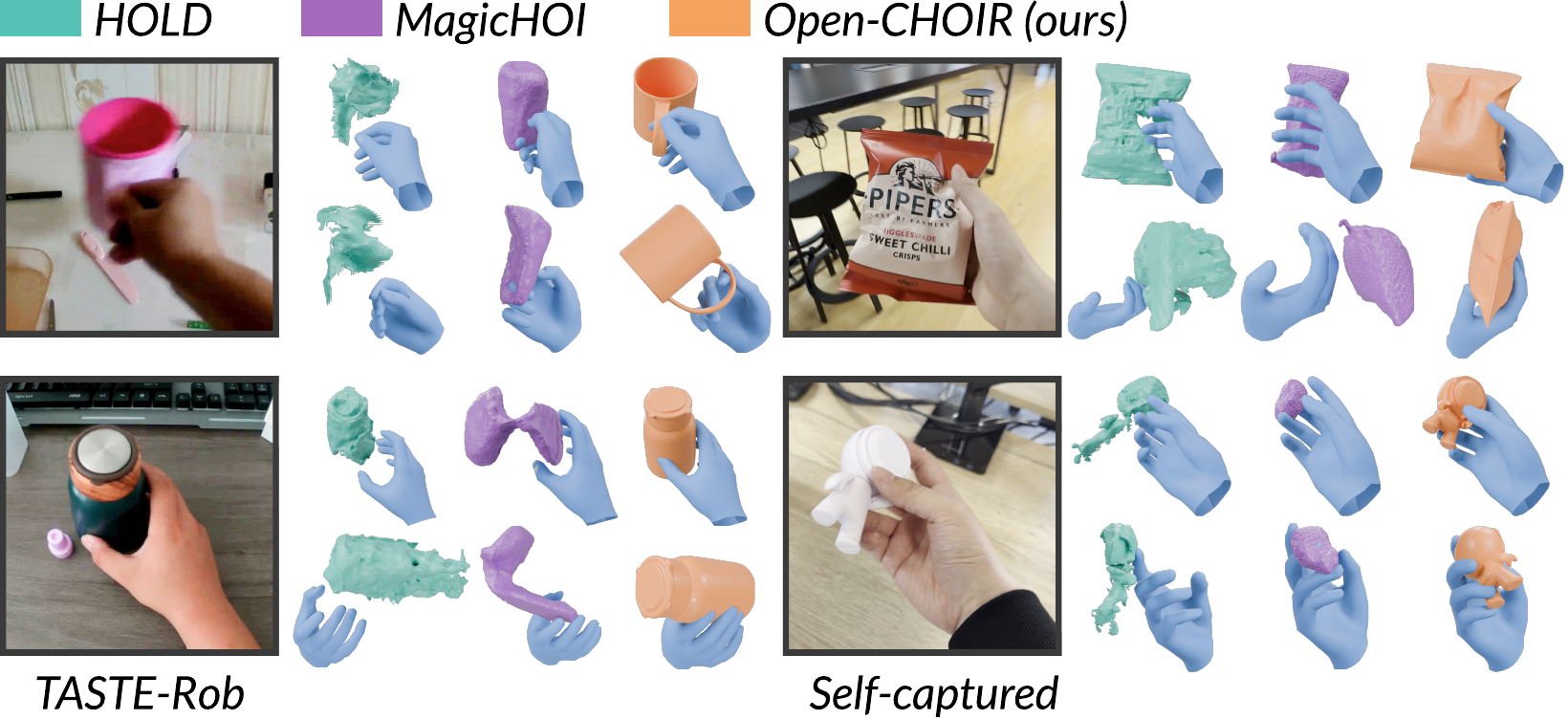}
    \vspace{-3pt}
    \caption{{{\color{black}Comparison with HOLD and MagicHOI on in-the-wild videos.}}}
    \label{fig:in_the_wild_baseline_comparison}
    \vspace{-5pt}
\end{figure}

\begin{figure*}[p]
    \centering
    \includegraphics[width=.95\linewidth]{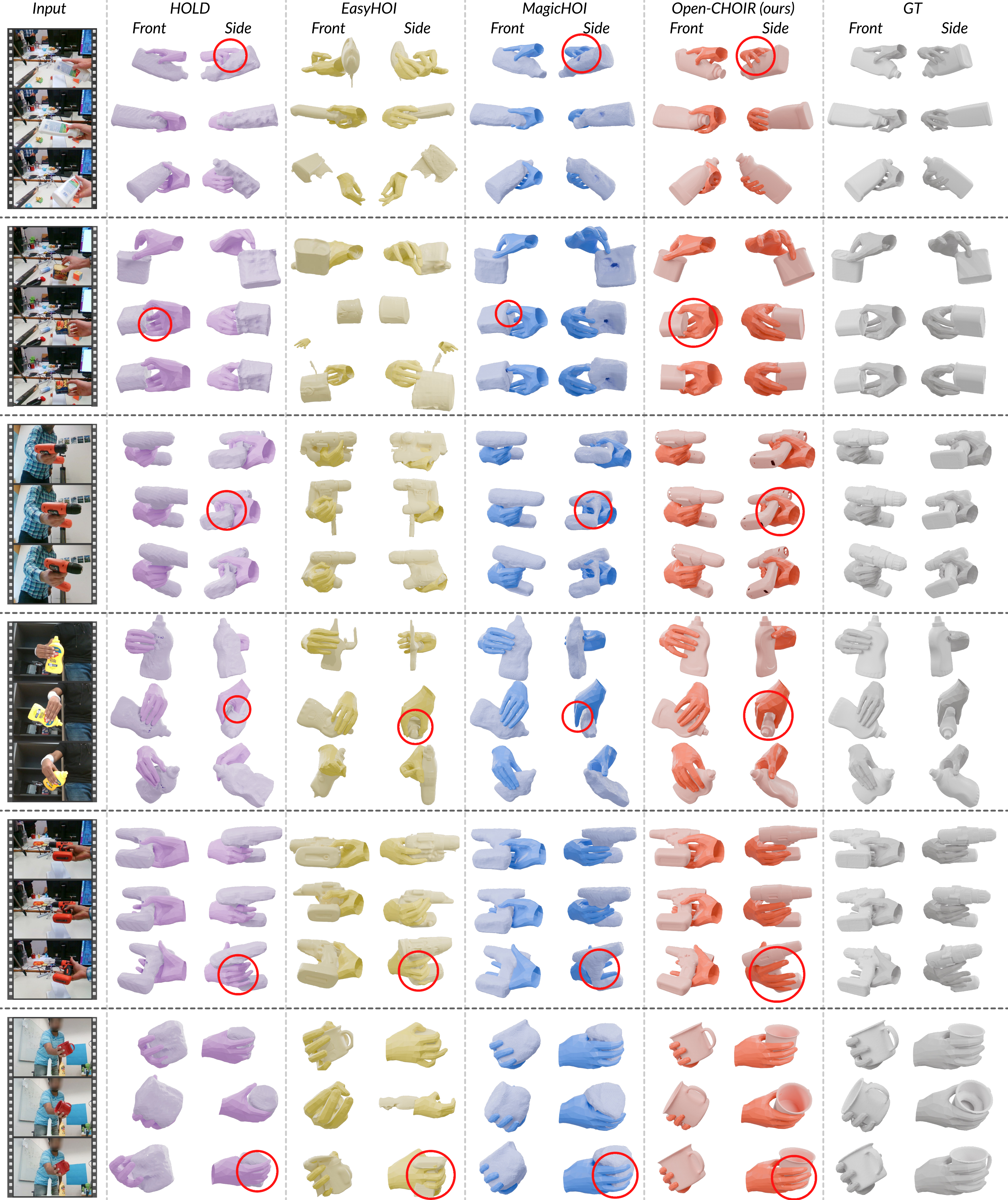}
    \vspace{-3pt}
    \caption{
        {{\color{black}Qualitative comparison between \name and state-of-the-art methods on HO3D. Red circles indicate regions of interest for comparing baseline reconstructions with ours. See videos on the supplementary webpage for temporal comparisons.}}
    }
    \label{fig:qualitative_ho3d}
    \vspace{-5pt}
\end{figure*}

\begin{figure*}[t]
    \centering
    \includegraphics[width=\linewidth]{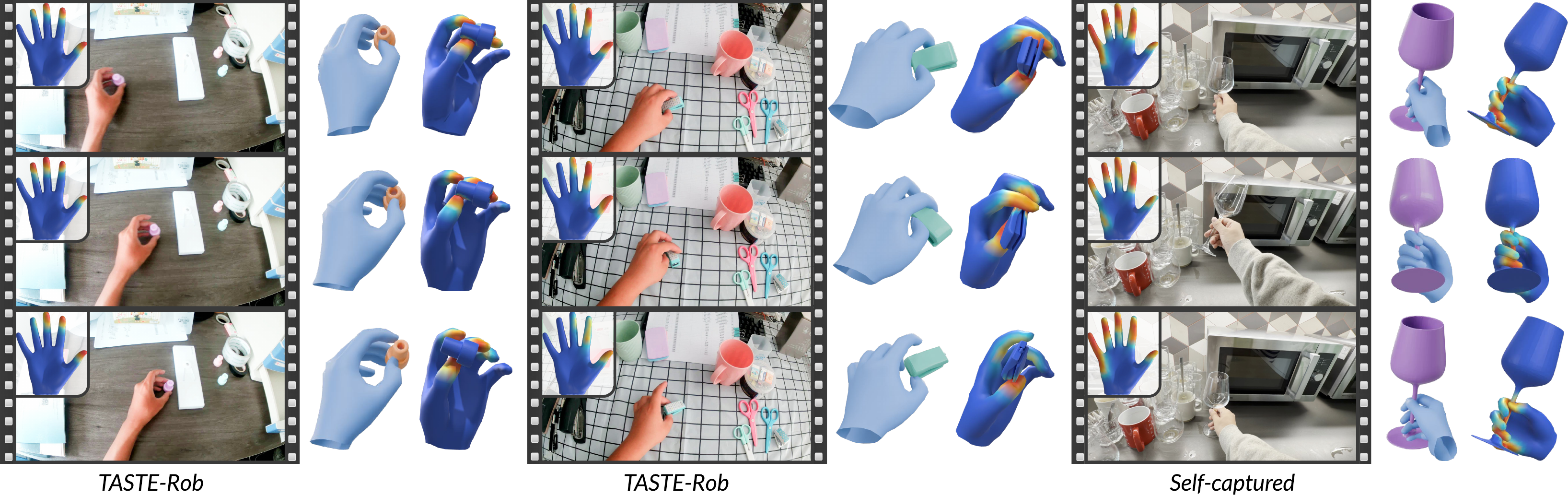}
    \vspace{-15pt}
    \caption{{Results of \name on TASTE-Rob and self-captured videos, showing input views, final Stage~3 novel-view geometry, and Stage~2 soft contact evidence (warmer colors indicate stronger guidance). See the supplementary webpage for temporal results.}}
    \label{fig:qualitative_taste_rob}
    \vspace{-3pt}
\end{figure*}

\paragraph{Qualitative comparison}
{{\color{black}
We first compare \name with baseline methods on HO3D and in-the-wild videos.
\Cref{fig:qualitative_ho3d} shows HO3D comparisons, and \Cref{fig:in_the_wild_baseline_comparison} compares \name with HOLD and MagicHOI on TASTE-Rob and self-captured videos, where the baseline pipelines produce usable reconstructions.
Compared with baselines that often suffer from missing contact, object interpenetration, or floating hands, our contact-aware optimization produces tighter hand--object alignment and fewer physically implausible penetration artifacts.
\Cref{fig:qualitative_taste_rob} further highlights challenging in-the-wild cases handled by our method, including small objects in the 1st column, changing hand poses and contacts over time in the 2nd column, and transparent objects with small contact areas in the 3rd column.
{The heatmaps summarize soft contact evidence from Stage~2 correspondences and proximity/normal-compatibility cues. Warmer regions provide stronger guidance for Stage~3, which balances contact with penetration, silhouette, hand/object anchors, 2D hand-joint and anatomical constraints, and temporal smoothness; small residual gaps may therefore remain in the final geometry.}
The supplementary webpage provides multi-view video comparisons against HOLD and MagicHOI, together with temporal visualizations of the contact maps.
These visualizations show that \name maintains coherent hand-object geometry and contact over time.
}}

\paragraph{Open-world/In-the-wild discussion}
{{\color{black}We clarify the scope of ``open-world'' and ``in-the-wild'' in four aspects.
(i) For \textit{objects}, \name does not require a category-specific template or a known 3D shape; the object geometry is reconstructed from the video, thus allowing unseen everyday objects. 
(ii) For \textit{camera}, \name operates on monocular RGB videos without requiring a calibrated static setup, and reconstructs hand-object motion in a consistent camera/reconstruction frame. 
(iii) For \textit{object motion}, we recover a rigid 6D pose trajectory, including both rotation and translation, instead of only tracking image-space masks or object centers. 
(iv) For \textit{contact}, our formulation builds contact evidence on contact-bearing frames and updates soft contact constraints during optimization, so the active hand vertices can vary during the interaction. 
However, we do not handle re-grasping or non-rigid deformations.}}

\begin{figure*}[t]
    \centering
    \includegraphics[width=\linewidth]{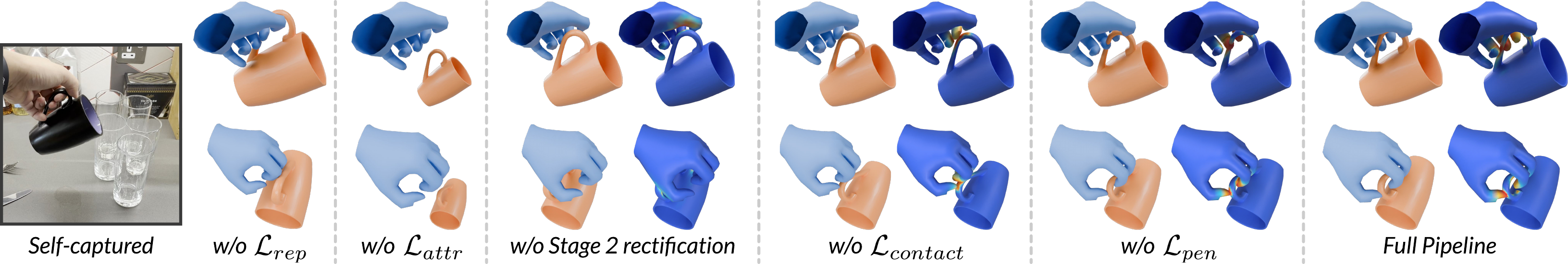}
    \vspace{-15pt}
    \caption{{{\color{black}Qualitative ablation of \name components against the full pipeline. The top and bottom rows show camera and side views, respectively.}}}
    \label{fig:wo_stage2_taste_rob}
    \vspace{-6pt}
\end{figure*}

\paragraph{Ablation studies}
{{\color{black}
We conduct ablation studies on the in-the-wild benchmark in \Cref{tab:ablation_in_the_wild}, where reference-free metrics directly measure visual alignment, hand-object proximity, penetration, and temporal smoothness.
We evaluate the cumulative contribution of all three stages on the full benchmark. 
%
%
Stage 1 achieves $77.86$ mIoU with a $0.31\,\mathrm{cm}$ gap and $4.46\%$ penetration. Stage 1+2 reduces these to $0.23\,\mathrm{cm}$ and $3.73\%$ at $77.29$ mIoU, while the full pipeline reaches a $0.10\,\mathrm{cm}$ gap and $76.65$ mIoU. Stage 3 also reduces hand/object acceleration from $0.45/0.15$ to $0.42/0.12\,\mathrm{cm/frame}^{2}$. Thus, Stage 2 corrects coarse placement and Stage 3 provides the strongest contact refinement.
Rows~(a) and~(b) show that the Stage~1 terms $\mathcal{L}_{\mathrm{rep}}$ and $\mathcal{L}_{\mathrm{attr}}$ are important for isolated object fitting: removing them substantially reduces mIoU and increases hand-object distance.
Row~(c) removes Stage~2 rectification; although the silhouette score remains similar, the hand-object distance roughly doubles, confirming that this stage mainly corrects relative placement.
Rows~(d)--(f) isolate the Stage~3 penetration, contact, and temporal terms, showing their effects on penetration handling, contact alignment, and motion smoothness.
{Rows~(a)--(f) are interpreted relative to the full model using the metrics jointly, as lower penetration reflects larger hand--object separation.}
\Cref{fig:wo_stage2_taste_rob} shows that removing key components leads to inferior object poses, missing contacts, or larger penetration artifacts.
\Cref{fig:contact} further shows that Stage~2 rectification improves the hand-object alignment before contact correspondences are built.
}}

\begin{table}[t]
    \centering
    \caption{{Ablation study on the in-the-wild benchmark. Rows~(a)--(f) remove individual components from the full pipeline.}}
    \vspace{-6pt}
    \label{tab:ablation_in_the_wild}
    \begingroup
        \footnotesize
        \setlength\tabcolsep{4pt}
        \begin{tabular*}{\linewidth}{@{\extracolsep{\fill}}lccccc@{}}
            \toprule
            \multirow{2}{*}{Method} & mIoU & Pen. Ratio & H-O Dist. & Acc$_h$ & Acc$_o$ \\
            \cmidrule(lr){2-6}
             & [\%] $\uparrow$ & [\%] $\downarrow$ & [cm] $\downarrow$ & [cm/fr$^2$] $\downarrow$ & [cm/fr$^2$] $\downarrow$ \\
            \midrule
            (a) Stage~1 w/o $\mathcal{L}_{\mathrm{rep}}$     & 59.24 & 7.82 & 0.13 & 0.43 & 0.09 \\
            (b) Stage~1 w/o $\mathcal{L}_{\mathrm{attr}}$    & 53.06 & 2.50 & 0.42 & 0.45 & 0.08 \\
            (c) w/o Stage~2 rect.              & 76.78 & 5.30 & 0.22 & 0.36 & 0.12 \\
            (d) Stage~3 w/o $\mathcal{L}_{\mathrm{pen}}$ & 76.65 & 6.37 & 0.10 & 0.42 & 0.12 \\
            (e) Stage~3 w/o $\mathcal{L}_{\mathrm{contact}}$ & 76.65 & 4.61 & 0.12 & 0.42 & 0.12 \\
            (f) Stage~3 w/o $\mathcal{L}_{\mathrm{temp}}$ & 76.65 & 5.90 & 0.11 & 0.43 & 0.12 \\
            \midrule
            (g) \name (ours)               & 76.65 & 5.81 & 0.10 & 0.42 & 0.12 \\
            \bottomrule
        \end{tabular*}
    \endgroup
    \vspace{-20pt}
\end{table}

\section{Conclusion}

We presented \name, an end-to-end framework for \emph{open-world}
4D hand--object interaction reconstruction from monocular videos. Our central enabler is a \textit{contact-first} formulation: we model contact as the transferable quantity to couple hand articulation and object geometry, allowing robust joint recovery even when initial monocular estimates are noisy and occluded. Combining {{\color{black}generative HOI spatial rectification}} with contact-aware test-time optimization, we produce temporally-consistent 3D hand motion, object shape, and 6D pose trajectories, as well as explicit grasp/release events, and generalize to unseen objects and cluttered real-world scenes.

\begin{figure}[H]
    \centering
    \includegraphics[width=\linewidth]{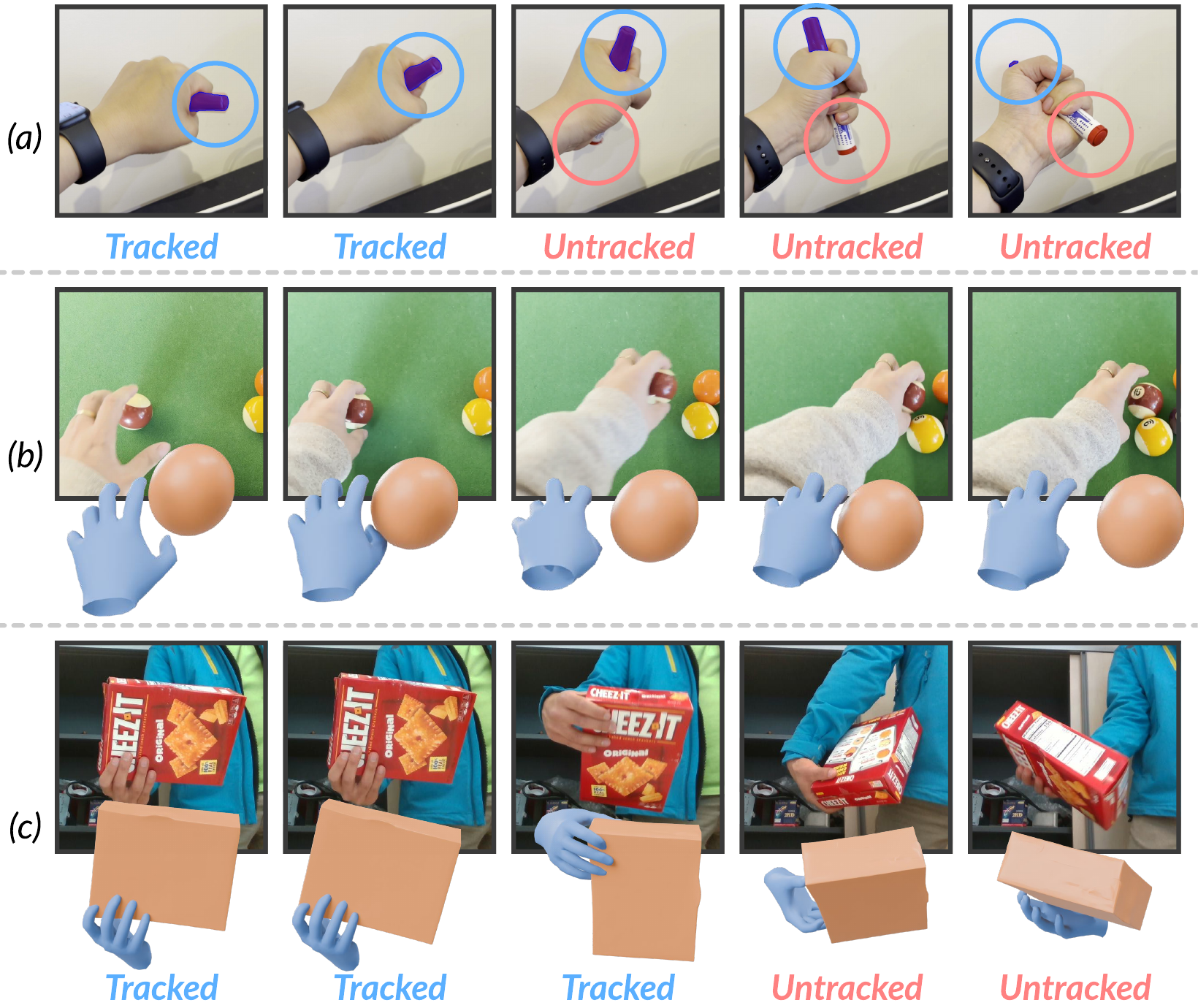}
    \vspace{-14pt}
    \caption{
        Representative failure cases: (a) incomplete object detection or segmentation, (b) inaccurate metric scale, and (c) ambiguous tracking of symmetric objects under rapid motion.
    }
    \label{fig:failure_cases}
    \vspace{-13pt}
\end{figure}

\vspace{8pt}
\paragraph{Limitations and future work} 
%
The main failures stem from Stage 1 object segmentation, scale estimation, and symmetric-object tracking (\Cref{fig:failure_cases}); later stages can improve local alignment but cannot correct these upstream errors. Incomplete masks may distort object geometry, while scale and pose ambiguities can preserve an incorrect interaction mode. The method also requires valid anchor frames, handles only single-hand interactions, and degrades under endpoint occlusion. The current formulation does not model re-grasping or non-rigid object deformation. Future work will exploit multi-view cues, support bimanual interactions, and replace iterative optimization with a feed-forward model.

\begin{acks}
This work was partially supported by the Research Grants Council of the Hong Kong Special Administrative Region, China, under Project T45-401/22-N, the CUHK Direct Grant for Research under Project 4055296, and gifts from Adobe and the UCL AI Centre.
\end{acks}

\clearpage
\section*{Supplementary Material}
\section{Implementation Details}

This supplementary material provides implementation details for the three stages of \name, focusing on thresholds, optimization terms, schedules, and engineering choices that are omitted from the main paper due to page limit.

\subsection{Stage 1: Open-World HOI Analysis}
\label{supp:stage1}

\paragraph{2D cue extraction.}
Given an RGB video, we estimate per-frame hand boxes, hand chirality, 2D hand joints, interacting-object boxes, and hand/object masks. The object pseudo-labels used for detector adaptation are mined around a motion-selected interaction frame. We identify this frame as the first local minimum in the fingertip velocity profile, which typically corresponds to the transition into grasping or manipulation. At this frame, we prompt SAM~2~\cite{ravi2024sam} with a fingertip-centered point grid (10$\times$10) and collect candidate object masks. We keep candidates whose temporal motion is correlated with the hand trajectory, producing object pseudo-labels tied to the manipulated object rather than background distractors. These pseudo-labels are then used to fine-tune WiLoR~\cite{potamias2025wilor} with an interacting-object box head.

\paragraph{Mask propagation and amodal refinement.}
At test time, we initialize the object mask from the frame with the highest object-detection confidence and propagate it bidirectionally through the video using SAM~2~\cite{ravi2024sam}. Because propagated masks are modal and may miss hand-occluded regions, we refine them with an amodal video segmenter~\cite{chen2025using}. The resulting amodal silhouettes are used by the object pose tracker and by the Stage~1 and Stage~3 silhouette losses.

\paragraph{Hand and object initialization.}
HaMeR~\cite{pavlakos2024reconstructing} initializes MANO~\cite{mano} hand parameters in each frame. VIPE~\cite{huang2025vipe} estimates camera extrinsics, and Dyn-HaMR~\cite{yu2025dyn} temporally stabilizes the hand trajectory. For the object, SAM-3D-Objects~\cite{sam3dteam2025sam3d3dfyimages} reconstructs a canonical mesh, metric scale, and an initial 6D pose from an object anchor frame, which defaults to the first frame but may be replaced by any frame with reliable object visibility. This anchor-frame reconstruction defines the fixed object geometry used by the follow tracker.

\paragraph{Guarded SAM-3D-Objects follow tracking.}
SAM-3D-Objects~\cite{sam3dteam2025sam3d3dfyimages} is a single-frame reconstructor. We therefore use it in a guarded follow-tracking procedure to obtain a dense object pose sequence. First, on the object anchor frame, we run full single-frame reconstruction and store the inferred object state. For subsequent frames, we initialize from this anchor-frame state, keep the shape, scale, and translation-scale latents fixed, and update only the rotation and translation pose latents. This preserves object geometry and metric scale while allowing each frame to re-estimate pose from visible image evidence. Optionally, we apply a weak pose-key velocity regularizer during follow inference:
\begin{equation}
\label{supp:eq_sam3d_guidance}
\mathbf{v}_t^{(R^o,T^o)}
\leftarrow
\mathbf{v}_t^{(R^o,T^o)}
-
\lambda_{\mathrm{temp}}
\left(
\mathbf{x}_t^{(R^o,T^o)}
-
\hat{\mathbf{x}}_{t,\mathrm{prev}}^{(R^o,T^o)}
\right),
\end{equation}
Here $\mathbf{x}_t^{(R^o,T^o)}$ denotes the current rotation/translation pose latent and $\hat{\mathbf{x}}_{t,\mathrm{prev}}^{(R^o,T^o)}$ denotes the propagated pose-key reference, with $\lambda_{\mathrm{temp}}=0.5$.
The primary safeguard is an angular guard: a candidate frame is rejected if its quaternion angular distance from the last accepted tracking estimate exceeds $60^\circ$. We retry with incremented random seeds up to three times; if all attempts fail, the frame is left missing and the last accepted estimate is not updated. Missing rotations are filled by spherical linear interpolation (SLERP) and translations by linear interpolation. The resulting dense object poses initialize Stage~1 isolated sequence fitting (Step 4).

\paragraph{Static object initialization.}
Before full-sequence fitting, we refine a shared object pose and scale on the visible static segment when available. The objective combines a soft silhouette mask loss and a metric depth loss:
\begin{equation}
\mathcal{L}_{\mathrm{static}}
=
\mathcal{L}_{\mathrm{mask}}
+
\mathcal{L}_{\mathrm{depth}} .
\end{equation}
Here $\mathcal{L}_{\mathrm{mask}}$ is the MSE between the rendered silhouette and the modal object mask, and $\mathcal{L}_{\mathrm{depth}}$ is the MSE between rendered depth and the MoGe-2 metric-depth estimate used by SAM-3D-Objects, evaluated on pixels that are both rendered and inside the modal mask.
The two terms are EMA-normalized and assigned equal relative weights. We optimize for 500 iterations with Adam~\cite{kingma2014adam} at learning rate $10^{-3}$, followed by an ICP sanity check accepted only when the induced scale change lies in $[0.7,1.3]$ and the amodal-mask IoU remains at least $0.8$.

\paragraph{Isolated sequence fitting.}
Stage~1 isolated fitting refines the object trajectory and MANO~\cite{mano} hand parameters before Stage~2 HOI spatial rectification. It uses the main-paper repulsion and attraction terms for object silhouette fitting, together with temporal/static object regularization and hand pose regularization.
For the object, we optimize
\begin{equation}
\begin{aligned}
\mathcal{L}_{o}^{\mathrm{iso}}
&=
\mathcal{L}_{\mathrm{rep}}
+
\mathcal{L}_{\mathrm{attr}}
+
\mathcal{L}_{\mathrm{temp}}^{o}
+
\mathcal{L}_{\mathrm{stat}}^{o}.
\end{aligned}
\end{equation}
We EMA-normalize the repulsion and attraction terms in implementation, and use weights $1.5$, $1.0$, $10$, and $1$ for $\mathcal{L}_{\mathrm{rep}}$, $\mathcal{L}_{\mathrm{attr}}$, $\mathcal{L}_{\mathrm{temp}}^{o}$, and $\mathcal{L}_{\mathrm{stat}}^{o}$, respectively. The object temporal term is
\begin{equation}
\mathcal{L}_{\mathrm{temp}}^{o}
=
\mathcal{L}_{\mathrm{sm\text{-}rot}}
+
\mathcal{L}_{\mathrm{sm\text{-}tr}} ,
\end{equation}
where $\mathcal{L}_{\mathrm{sm\text{-}rot}}$ and $\mathcal{L}_{\mathrm{sm\text{-}tr}}$ impose temporal smoothness on object rotation and translation.
Let $\mathbf{M}$ denote the rendered object silhouette, $\hat{\mathbf{M}}$ the target amodal object mask, and $\Omega$ the image domain. The repulsion loss penalizes rendered pixels outside the target mask:
\begin{equation}
\mathcal{L}_{\mathrm{rep}}
=
\frac{1}{|\Omega|}
\sum_{\mathbf{p}\in\Omega}
\left[
\max\left(\mathbf{M}_{\mathbf{p}}-\hat{\mathbf{M}}_{\mathbf{p}},0\right)
\right]^2 .
\end{equation}
The attraction loss draws $2{,}000$ samples from the target-mask region and gives extra probability to target pixels not yet covered by the rendered silhouette. We sample pixels from
\begin{equation}
\mathcal{P}(\mathbf{p})
\propto
\hat{\mathbf{M}}_{\mathbf{p}}
+
\alpha_{\mathrm{uncover}}\,
\hat{\mathbf{M}}_{\mathbf{p}}(1-\mathbf{M}_{\mathbf{p}})
+
\varepsilon ,
\end{equation}
where $\alpha_{\mathrm{uncover}}=20$. The sampled set $\mathcal{S}$ is restricted to target-mask pixels and is biased toward uncovered regions. Consistent with the main paper, we then define
\begin{equation}
\mathcal{L}_{\mathrm{attr}}
=
\frac{1}{|\mathcal{S}|}
\sum_{\mathbf{p}\in\mathcal{S}}
\min_{\mathbf{v}\in\mathcal{V}}\|\mathbf{p}-\Pi(\mathbf{v})\|_2^2 .
\end{equation}
Here $\mathcal{S}$ is the sampled pixel set, $\mathcal{V}$ is the object-vertex set, and $\Pi(\cdot)$ is the camera projection. The static-consistency term ties pre- and post-interaction static segments to segment-level average object poses:
\begin{equation}
\mathcal{L}_{\mathrm{stat}}^{o}
=
\sum_{S\in\{\mathrm{pre},\mathrm{post}\}}
\left(
\|\mathbf{R}^{o}_{S}-\bar{\mathbf{R}}^{o}_{S}\|_2^2
+
\|\mathbf{T}^{o}_{S}-\bar{\mathbf{T}}^{o}_{S}\|_2^2
\right),
\end{equation}
where $\mathbf{R}^{o}_{S}$ and $\mathbf{T}^{o}_{S}$ are the object rotation and translation for static segment $S$, and $\bar{\mathbf{R}}^{o}_{S}$ and $\bar{\mathbf{T}}^{o}_{S}$ denote their segment averages. This static consistency is assigned an inline weight of $10^2$.

For the hand, we optimize
\begin{equation}
\begin{aligned}
\mathcal{L}_{h}^{\mathrm{iso}}
&=
\mathcal{L}_{2D}^{h}
+
\mathcal{L}_{\mathrm{anat}}^{h}
+
\mathcal{L}_{\mathrm{prior}}^{h}
+
\mathcal{L}_{\mathrm{depth}}^{h}
+
\mathcal{L}_{\mathrm{temp}}^{h}.
\end{aligned}
\end{equation}
Here $\mathcal{L}_{2D}^{h}$ is the 2D joint reprojection loss, $\mathcal{L}_{\mathrm{anat}}^{h}$ is the MANO~\cite{mano} twist-splay-bend anatomy constraint, $\mathcal{L}_{\mathrm{prior}}^{h}$ anchors the hand to the HaMeR~\cite{pavlakos2024reconstructing} MANO pose in 6D rotation space, $\mathcal{L}_{\mathrm{depth}}^{h}$ anchors wrist depth to the median metric-depth estimate inside the hand mask, and $\mathcal{L}_{\mathrm{temp}}^{h}$ is first-order smoothness on 3D hand joints. We EMA-normalize $\mathcal{L}_{2D}^{h}$ and use unit weight for it after normalization. We use weights $5$, $1$, $10$, and $\max(0.1\lambda_{\mathrm{temp}}^{o},1.0)$ for $\mathcal{L}_{\mathrm{anat}}^{h}$, $\mathcal{L}_{\mathrm{prior}}^{h}$, $\mathcal{L}_{\mathrm{depth}}^{h}$, and $\mathcal{L}_{\mathrm{temp}}^{h}$, respectively.
After isolated fitting, ray-scale alignment shifts the object trajectory along the camera ray so that its interaction-depth statistics better match the hand sequence, producing the coarse contact-agnostic 4D HOI sequence used by Stage~2.

\subsection{Stage 2: Generative HOI Spatial Rectification}
\label{supp:stage2}

\paragraph{Generative HOI spatial rectification architecture.}
The rectification network follows the architecture illustrated in the main paper. Training pairs are derived from DexGraspNet~\cite{wang2022dexgraspnet} and filtered with PyBullet~\cite{coumans2016pybullet}; 
we uniformly sample object surface points $\mathbf{P}^{o}\in\mathbb{R}^{N\times3}$ and normals $\mathbf{n}^{o}$, normalize the object to a unit sphere, and build dense geometry tokens from $N'$ farthest-downsampled points using a lightweight point backbone inspired by PointNet++~\cite{qi2017pointnet++}. The encoder is coordinate-sensitive rather than rotation-invariant, because contact validity depends on absolute object orientation.

At flow-matching time $\tau$~\cite{lipman2022flow}, the scalar state $z_\tau$ represents the depth offset along the camera ray. We first embed the time, state, noisy hand joints, object scale, and viewing ray as $\mathbf{e}_{\tau}$, $\mathbf{e}_{z}$, $\mathbf{e}_{J}$, $\mathbf{e}_{s}$, and $\mathbf{e}_{r}$ using token-specific MLPs. The state and condition tokens are then formed as
\begin{equation}
\begin{gathered}
\mathbf{H}_{0}
=
\operatorname{Concat}\!\left(
\mathbf{e}_{\tau},
\mathbf{e}_{z},
\mathbf{e}_{J},
\mathbf{e}_{s},
\mathbf{e}_{r}
\right)
+\mathbf{E}_{\mathrm{PE}},\\
\mathbf{H}_{\mathrm{self}}
=
\operatorname{Attn}(\mathbf{H}_{0},\mathbf{H}_{0},\mathbf{H}_{0}).
\end{gathered}
\end{equation}
We obtain object geometry tokens using a PointNet++-style point encoder~\cite{qi2017pointnet++}, denoted by $\mathcal{E}$:
\begin{equation}
\mathbf{H}_{\mathrm{obj}}
=
\mathcal{E}(\mathbf{P}^{o},\mathbf{n}^{o})
+\mathbf{E}^{o}_{\mathrm{PE}},
\end{equation}
and the state/condition tokens query them through cross-attention:
\begin{equation}
\mathbf{H}_{\mathrm{cross}}
=
\operatorname{Attn}(\mathbf{H}_{\mathrm{self}},\mathbf{H}_{\mathrm{obj}},\mathbf{H}_{\mathrm{obj}}).
\end{equation}
Here $\operatorname{Attn}(\mathbf{Q},\mathbf{K},\mathbf{V})=\operatorname{softmax}(\mathbf{Q}\mathbf{K}^{\top}/\sqrt{d})\mathbf{V}$ and $\mathbf{E}_{\mathrm{PE}}$ denotes positional embedding, following standard Transformer~\cite{vaswani2017attention}. The flow-matching time $\tau$ is also injected into Transformer blocks via AdaLN-style conditioning~\cite{peebles2023scalable}. The token corresponding to $z_\tau$ is decoded into the scalar velocity $v_\theta(z_\tau,\tau \mid \mathbf{J}^{h},\mathbf{r},s,\mathbf{P}^{o},\mathbf{n}^{o})$.

\paragraph{Initial anchor construction.}
After applying the generative HOI ray-depth correction to each interaction frame, we construct initial contact correspondences on the rectified hand-object geometry. We align each per-frame object mesh to the first interaction-frame canonical mesh, sample 10,000 canonical object-surface points with a fixed seed, and search the $K=50$ nearest samples for each candidate MANO~\cite{mano} contact vertex. A correspondence is kept only if its distance is below $2\,\mathrm{cm}$ and the hand-normal/object-direction angle lies within a $60^\circ$ cone. The accepted point is projected back to an object face and stored as face index plus barycentric coordinates.

\subsection{Stage 3: Contact-aware HOI Optimization}
\label{supp:stage3}

\paragraph{Phase segmentation.}
Each video is divided into pre-static, approach, manipulation, release, and post-static phases when present. The segmentation combines object point-trajectory motion from CoTracker~\cite{karaev2025cotracker3} with frame-to-frame IoU changes of the amodal object masks. Continuous-contact clips may contain only the manipulation phase; missing phases are skipped rather than forced.

\paragraph{Objective expansion.}
We detail the five loss families used in Stage~3: contact $\mathcal{L}_{\mathrm{con}}$, penetration $\mathcal{L}_{\mathrm{pen}}$, silhouette $\mathcal{L}_{\mathrm{sil}}$, anchor $\mathcal{L}_{\mathrm{anc}}$, and temporal $\mathcal{L}_{\mathrm{temp}}$.
These terms are applied as test-time reconstruction constraints over a 4D sequence.

In implementation, the contact family contains the soft correspondence term and lightweight cache-stabilization terms:
\begin{equation}
\mathcal{L}_{\mathrm{con}}
=
\mathcal{L}_{\mathrm{contact}}
+
\mathcal{L}_{\mathrm{template}}
+
\mathcal{L}_{\mathrm{bridge}}
+
\mathcal{L}_{\mathrm{gap}}
+
\mathcal{L}_{\mathrm{patch}} .
\end{equation}
$\mathcal{L}_{\mathrm{contact}}$ is the soft correspondence loss that pulls active hand contact vertices toward their barycentric object anchors.
$\mathcal{L}_{\mathrm{template}}$, $\mathcal{L}_{\mathrm{bridge}}$, $\mathcal{L}_{\mathrm{gap}}$, and $\mathcal{L}_{\mathrm{patch}}$ stabilize reliable contacts across short temporal gaps and local neighborhoods. We use weight $10^3$ for $\mathcal{L}_{\mathrm{contact}}$; the auxiliary stabilizers use annealed weights and are included in the released code.

The soft contact term is restricted to interaction frames:
\begin{equation}
\mathcal{L}_{\mathrm{contact}}
=
\frac{
\sum_{t,i}
c_{t,i}
\sum_{k=1}^{8}
\omega_{t,i,k}
\|\mathbf{v}^{h}_{t,i}-\mathbf{a}^{o}_{t,i,k}\|_2^2
}{
\sum_{t,i} c_{t,i}
}.
\end{equation}
Here $\mathbf{a}^{o}_{t,i,k}$ are barycentric object anchors, $\omega_{t,i,k}$ are softmax weights over the top-$8$ anchors computed from surface-sample distances with $\sigma=0.01$, and $c_{t,i}$ is the Stage~3 contact confidence from current proximity/normal evidence and temporal memory. The soft correspondence cache is periodically rebuilt from the current optimized geometry, seeded by Stage~2 rectified contact evidence. Each rebuild uses a $5\,\mathrm{cm}$ distance threshold and a $60^\circ$ normal-compatibility cone, with temporal memory for stable correspondences.

The penetration term is applied to all frames and is one-sided:
\begin{equation}
\mathcal{L}_{\mathrm{pen}}
=
\frac{1}{T|\mathcal{V}^{h}|}
\sum_t
\sum_{\mathbf{v}^{h}\in\mathcal{V}^{h}_t}
\min\bigl(
\max(
(\mathbf{v}^{o}_{\mathrm{nn}}-\mathbf{v}^h)^\top
\mathbf{n}^{o}_{\mathrm{nn}}-\epsilon_{\mathrm{pen}},0),
0.04
\bigr),
\end{equation}
where $\mathcal{V}^{h}_t$ is the hand-vertex set at frame $t$; $\mathbf{v}^{o}_{\mathrm{nn}}$ and $\mathbf{n}^{o}_{\mathrm{nn}}$ denote the nearest object vertex and its outward normal for the hand vertex $\mathbf{v}^{h}$,
with dead zone $\epsilon_{\mathrm{pen}}=5\,\mathrm{mm}$, per-vertex residual clipping threshold $0.04$, and inline weight $500$.

The anchor family is decomposed as
\begin{equation}
\mathcal{L}_{\mathrm{anc}}
=
\mathcal{L}_{2D}^{h}
+
\mathcal{L}_{\mathrm{anat}}^{h}
+
\mathcal{L}_{\mathrm{pose}}^{h}
+
\mathcal{L}_{\mathrm{pose}}^{o} .
\end{equation}
We use weights $0.5$, $30$, and $100$ for $\mathcal{L}_{2D}^{h}$, $\mathcal{L}_{\mathrm{anat}}^{h}$, and $\mathcal{L}_{\mathrm{pose}}^{h}$, respectively. The object pose-anchor term is split as
\begin{equation}
\mathcal{L}_{\mathrm{pose}}^{o}
=
\mathcal{L}_{\mathrm{obj\text{-}anchor}}^{\mathrm{non\text{-}inter}}
+
\mathcal{L}_{\mathrm{obj\text{-}anchor}}^{\mathrm{inter}},
\end{equation}
with inline weight $100$ on each component. 
The non-interaction component anchors object poses in pre-/post-static or non-contact frames to the Stage~1 isolated trajectory, while the interaction component anchors interaction-frame object poses to the Stage~2 rectified initialization. This prevents contact forces from drifting the object away from image evidence.
The silhouette group $\mathcal{L}_{\mathrm{sil}}$ aligns the rendered object silhouette with the amodal object mask and uses inline weight $500$.

The temporal group is
\begin{equation}
\begin{aligned}
\mathcal{L}_{\mathrm{temp}}
&=
\mathcal{L}_{\mathrm{pose\text{-}vel}}
+
\mathcal{L}_{\mathrm{obj\text{-}vel}}
+
\mathcal{L}_{\mathrm{wrist\text{-}anchor}}
\\
&\quad+
\mathcal{L}_{\mathrm{hand\text{-}tr\text{-}vel}}
+
\mathcal{L}_{\mathrm{root\text{-}R\text{-}vel}}
+
\mathcal{L}_{\mathrm{pose\text{-}acc}}
\\
&\quad+
\mathcal{L}_{\mathrm{hand\text{-}tr\text{-}acc}}
+
\mathcal{L}_{\mathrm{root\text{-}R\text{-}acc}} .
\end{aligned}
\end{equation}
$\mathcal{L}_{\mathrm{wrist\text{-}anchor}}$ anchors wrist translation to the rectified/input trajectory. Subscripts ``vel'' and ``acc'' denote first- and second-order temporal finite differences, while ``R'' and ``tr'' denote rotation and translation.
The corresponding inline weights are $500$, $500$, $200$, $200$, $500$, $1{,}000$, $5{,}000$, and $3{,}000$ in the order written above. MANO~\cite{mano} finger-pose and root-rotation smoothness are computed in 6D-rotation or rotation-matrix space to avoid axis-angle wrap-around artifacts. Stage~3 runs for 800 iterations with Adam~\cite{kingma2014adam} using learning rates $3{\times}10^{-4}$ for object pose, $5{\times}10^{-4}$ for MANO finger pose, and $5{\times}10^{-5}$ for wrist rotation/translation.

\section{Evaluation Details}

\paragraph{Evaluation set statistics}
Table~\ref{tab:supp_eval_set_statistics} summarizes the number of available and evaluated cases for each dataset and method. The in-the-wild benchmark contains 70 TASTE-Rob videos and 30 self-captured videos. We count a baseline sequence as successfully reconstructed when at least one third of its frames produce usable reconstructions; under this criterion, HOLD succeeds on 30 sequences and MagicHOI succeeds on 22 sequences, while \name is evaluated on all 100 videos.

\begin{table}[t]
    \centering
    \caption{Evaluation-set statistics. ``Total cases'' denotes the available cases, while method columns report the number of successfully evaluated cases.}
    \label{tab:supp_eval_set_statistics}
    \begingroup
        \footnotesize
        \begin{tabular*}{\linewidth}{@{\extracolsep{\fill}}lccccc@{}}
            \toprule
            Dataset & Total cases & 3D GT & HOLD & MagicHOI & \name \\
            \midrule
            HO3D & 14 & Yes & 14 & 14 & 14 \\
            TASTE-Rob & 70 & No & 22 & 14 & 70 \\
            Self-captured & 30 & No & 8 & 8 & 30 \\
            Total & 114 & -- & 44 & 36 & 114 \\
            \bottomrule
        \end{tabular*}
    \endgroup
\end{table}

\paragraph{Reference-free metric details}
For datasets without 3D ground truth, we report reference-free image alignment, physical plausibility, and temporal smoothness metrics. Let $\mathcal{T}$ be the set of frames with valid reconstructions for a method. For baselines with missing outputs, all per-frame metrics are averaged over $\mathcal{T}$, and temporal accelerations are computed only on consecutive valid frame triplets.

For video alignment, let $\mathbf{M}^{\mathrm{rend}}_t$ be the rendered hand-object silhouette and $\mathbf{M}^{\mathrm{obs}}_t$ be the tracked 2D hand-object mask at frame $t$. We compute
\begin{equation}
\mathrm{mIoU}
=
\frac{1}{|\mathcal{T}|}
\sum_{t\in\mathcal{T}}
\frac{
|\mathbf{M}^{\mathrm{rend}}_t \cap \mathbf{M}^{\mathrm{obs}}_t|
}{
|\mathbf{M}^{\mathrm{rend}}_t \cup \mathbf{M}^{\mathrm{obs}}_t|
}.
\end{equation}
For hand-object proximity, let $\mathcal{V}^{h}_t$ and $\mathcal{V}^{o}_t$ denote the reconstructed hand and object vertices in frame $t$. The hand-object distance is the closest surface distance,
\begin{equation}
\mathrm{H\text{-}O~Dist.}
=
\frac{1}{|\mathcal{T}|}
\sum_{t\in\mathcal{T}}
\min_{\mathbf{v}^{h}\in\mathcal{V}^{h}_t,\,
\mathbf{v}^{o}\in\mathcal{V}^{o}_t}
\|\mathbf{v}^{h}-\mathbf{v}^{o}\|_2 .
\end{equation}
For penetration, let $d^o_t(\mathbf{v}^{h})$ be the signed distance from a hand vertex to the object, with positive values inside the object and non-positive values outside. We report the percentage of hand vertices inside the object,
\begin{equation}
\mathrm{Pen.~Ratio}
=
\frac{100}{|\mathcal{T}|}
\sum_{t\in\mathcal{T}}
\frac{1}{|\mathcal{V}^{h}_t|}
\sum_{\mathbf{v}^{h}\in\mathcal{V}^{h}_t}
\mathbb{1}\!\left[d^o_t(\mathbf{v}^{h})>0\right].
\end{equation}
Finally, for temporal smoothness, let $\mathbf{J}^{h}_{t,j}$ be the 3D position of hand joint $j$ at frame $t$, with $J$ denoting the number of hand joints. We define the object center $\mathbf{c}^{o}_t$ as the centroid of the reconstructed object vertices $\mathcal{V}^{o}_t$, and let $\mathcal{A}\subset\mathcal{T}$ be the set of valid middle frames whose neighbors $t-1$ and $t+1$ are also valid. With positions measured in centimeters, we compute
\begin{equation}
\mathrm{Acc}_h
=
\frac{1}{|\mathcal{A}|J}
\sum_{t\in\mathcal{A}}
\sum_{j=1}^{J}
\left\|
\mathbf{J}^{h}_{t+1,j}
-2\mathbf{J}^{h}_{t,j}
+\mathbf{J}^{h}_{t-1,j}
\right\|_2 ,
\end{equation}
and use object centers as a common trajectory representation for object acceleration,
\begin{equation}
\mathrm{Acc}_o
=
\frac{1}{|\mathcal{A}|}
\sum_{t\in\mathcal{A}}
\left\|
\mathbf{c}^{o}_{t+1}
-2\mathbf{c}^{o}_{t}
+\mathbf{c}^{o}_{t-1}
\right\|_2 .
\end{equation}
Both acceleration metrics are reported in $\mathrm{cm/frame}^2$.

\section{Additional Results}

This section collects supplementary visual diagnostics, ablations, runtime details, and failure-case analysis. The visual diagnostics are included here because they expose the intermediate image evidence used by the later 4D reconstruction and contact-aware optimization stages.

\paragraph{Stage 1 2D assets}
\label{supp:stage1_2d_assets}

\begin{figure*}[p]
    \centering
    \includegraphics[width=\textwidth]{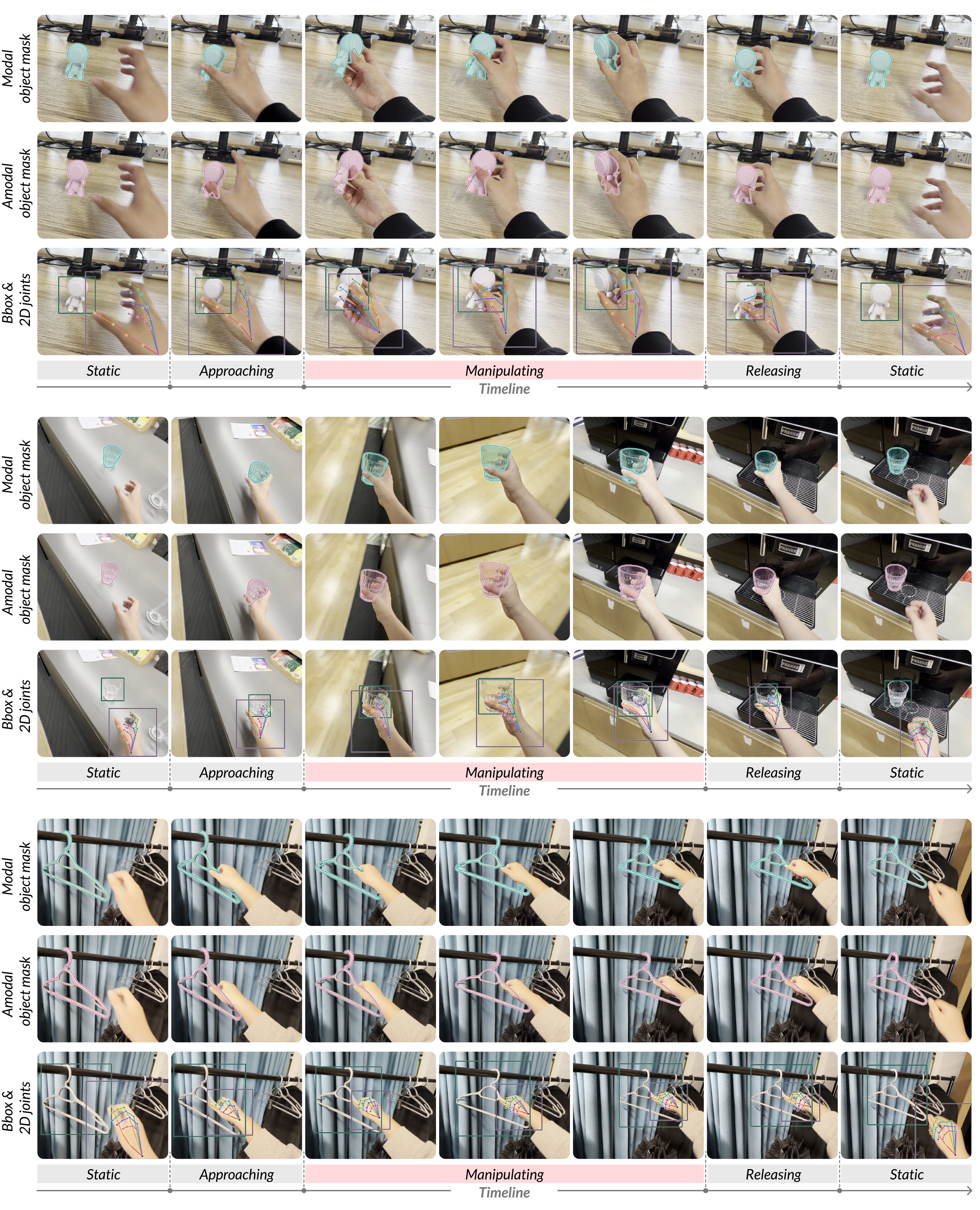}
    \caption{
    Visualization of Stage~1 2D assets. The figure shows modal object masks, amodal object masks, 2D hand and object bounding boxes, and 2D hand joints, which provide the image-space evidence used by the subsequent 3D reconstruction, HOI spatial rectification, and optimization stages.
    }
    \Description{Supplementary visualization of Stage~1 image-space outputs, including object masks, hand and object bounding boxes, and 2D hand joints.}
    \label{fig:supp_2d_assets}
\end{figure*}

Stage~1 produces modal masks, amodal masks, 2D hand/object boxes, and 2D hand joints as image-space evidence for the later 4D reconstruction stages. \Cref{fig:supp_2d_assets} shows representative outputs.

\paragraph{HO3D ablation}
In Table~\ref{tab:supp_ho3d_ablation}, we include the HO3D object-centric ablation for completeness. Several Stage~3 variants produce close object-centric scores because these metrics are less sensitive to contact-side finger updates than to object geometry, object pose, and hand-root/object placement.

\begin{table}[t]
    \centering
    \caption{Supplementary HO3D ablation study.}
    \label{tab:supp_ho3d_ablation}
    \begingroup
        \footnotesize
        \setlength\tabcolsep{1pt}
        \begin{tabular*}{\linewidth}{@{\extracolsep{\fill}}lcccccc@{}}
            \toprule
            \multirow{2}{*}{Method} & CD & F5 & F10 & MPJPE & CD$_h$ & RS \\
            \cmidrule(lr){2-7}
             & [cm] $\downarrow$ & [\%] $\uparrow$ & [\%] $\uparrow$ & [mm] $\downarrow$ & [cm] $\downarrow$ & $\downarrow$ \\
            \midrule
            (a) Stage~1 w/o $\mathcal{L}_{\mathrm{rep}}$ & 0.771 & 72.41 & 96.62 & 5.84 & 7.24 & 0.38 \\
            (b) Stage~1 w/o $\mathcal{L}_{\mathrm{attr}}$ & 0.833 & 69.88 & 94.79 & 5.76 & 5.65 & 0.25 \\
            (c) w/o Stage~2 rect. & 0.770 & 72.15 & 97.27 & 6.02 & 4.49 & 0.10 \\
            (d) Stage~3 w/o $\mathcal{L}_{\mathrm{pen}}$ & 0.770 & 73.26 & 95.96 & 5.55 & 2.41 & 0.10 \\
            (e) Stage~3 w/o $\mathcal{L}_{\mathrm{contact}}$ & 0.771 & 72.57 & 96.62 & 5.55 & 2.41 & 0.10 \\
            (f) Stage~3 w/o $\mathcal{L}_{\mathrm{temp}}$ & 0.768 & 72.75 & 96.90 & 5.55 & 2.41 & 0.10 \\
            \midrule
            (g) \textbf{\name (ours)} & \textbf{0.772} & \textbf{72.33} & \textbf{96.03} & \textbf{5.55} & \textbf{2.41} & \textbf{0.10} \\
            \bottomrule
        \end{tabular*}
    \endgroup
\end{table}

\paragraph{Bimanual extension cases}
Although the main pipeline is designed for single-hand interactions, it can be adapted with simple modifications to bimanual interactions where two hands manipulate different objects or the same object, as visualized on the supplementary webpage. When the two hands interact with different objects, we run the single-hand pipeline independently for each hand-object pair. In such cases, the hand scale and 2D joint constraints help reduce monocular depth ambiguity, so the two recovered interactions often remain visually compatible; however, there is no explicit cross-hand depth constraint or joint optimization, so global consistency is not strictly guaranteed.

For the case where two hands interact with the same object, we share the object geometry reconstructed from the same anchor frame. We then run Stages~2--3 separately for the left and right hand, obtaining two optimized hand-object sequences that use the same object shape. To merge them, we compute the rigid transformation between the two optimized object poses and use this relative transform to map one hand's global pose into the coordinate frame of the other hand-object reconstruction. This produces a common-coordinate visualization of two hands interacting with the same object. This procedure is a post-hoc composition for visualization rather than a full bimanual optimization, and future work should model both hands and their contacts jointly.

\paragraph{Runtime}
\label{supp:runtime}

All experiments are implemented in PyTorch~\cite{pytorch19} and run on NVIDIA RTX~4090 GPUs. We report runtime per video, averaged over clips with approximately 200 frames. The total inference time is around 10 hours for HOLD~\cite{fan2024hold}, around 9 hours for EasyHOI~\cite{liu2025easyhoi}, around 2 hours for MagicHOI~\cite{wang2025magichoi}, and around 30 minutes for \name. For our pipeline, Stage~1 takes around 20 minutes for 2D cue extraction and mask processing, around 5 minutes for object keyframe reconstruction and guarded follow tracking, and around 30 seconds for isolated sequence fitting. Stage~2 generative HOI spatial rectification takes around 10 seconds, and Stage~3 contact-aware optimization takes around 2 minutes. The one-time training cost of the generative HOI spatial rectification module is not included in per-video inference runtime.

\section{Limitations}
\label{supp:limitations}



\name assumes a single hand manipulating a rigid object in a monocular RGB video. It can handle clutter, occlusion, and unknown object categories, but may fail when object geometry cannot be recovered from the anchor frame, when the object is fully occluded for a long period, or when strong rotational symmetry makes the spin angle unobservable from silhouettes and contacts alone. Non-rigid objects, articulated objects, and full bimanual joint optimization are promising directions for future work.

\clearpage


\end{document}